# Penalized model-based clustering with cluster-specific diagonal covariance matrices and grouped variables

**Benhuai Xie and Wei Pan***

*Division of Biostatistics, School of Public Health, University of Minnesota*
*e-mail:* benhuaix@biostat.umn.edu; weip@biostat.umn.edu
*url:* www.biostat.umn.edu/∼weip

**Xiaotong Shen**

*School of Statistics, University of Minnesota*
*e-mail:* xshen@stat.umn.edu

**Abstract:** Clustering analysis is one of the most widely used statistical tools in many emerging areas such as microarray data analysis. For microarray and other high-dimensional data, the presence of many noise variables may mask underlying clustering structures. Hence removing noise variables via variable selection is necessary. For simultaneous variable selection and parameter estimation, existing penalized likelihood approaches in model-based clustering analysis all assume a common diagonal covariance matrix across clusters, which however may not hold in practice. To analyze high-dimensional data, particularly those with relatively low sample sizes, this article introduces a novel approach that shrinks the variances together with means, in a more general situation with cluster-specific (diagonal) covariance matrices. Furthermore, selection of grouped variables via inclusion or exclusion of a group of variables altogether is permitted by a specific form of penalty, which facilitates incorporating subject-matter knowledge, such as gene functions in clustering microarray samples for disease subtype discovery. For implementation, EM algorithms are derived for parameter estimation, in which the M-steps clearly demonstrate the effects of shrinkage and thresholding. Numerical examples, including an application to acute leukemia subtype discovery with microarray gene expression data, are provided to demonstrate the utility and advantage of the proposed method.

## Contents

*Corresponding author.

## 1. Introduction

Clustering analysis is perhaps the most widely used analysis method for microarray data: it has been used for gene function discovery (Eisen et al. 1998 [10]) and cancer subtype discovery (Golub et al. 1999 [15]). In such an application involving a large number of genes arrayed, it is necessary but challenging to choose a set of informative genes for clustering. If some informative ones are excluded because fewer genes are used, then it becomes difficult or impossible to discriminate some phenotypes of interest such as cancer subtypes. On the other hand, using redundant genes introduces noise, leading to the failure to uncover the underlying clustering structure. For example, Alaiya et al. (2002) [1] considered borderline ovarian tumor classification via clustering protein expression profiles: using all 1584 protein spots on an array failed to achieve an accurate classification, while an appropriate selection of spots (based on discriminating between benign and malignant tumors) did give biologically more meaningful results.

In spite of its importance, it is not always clear how to select genes for clustering. In particular, as demonstrated by Pan and Shen (2007) [39] and Pan et al. (2006) [37], unlike in the context of supervised learning, including regression, best subset selection, one of the most widely used model selection methods for supervised learning, fails for clustering and semi-supervised learning, in addition to its prohibitive computing cost for high-dimensional data; the reason is the existence of many correct models, most of which are not of interest. In a review of the earlier literature on this problem, Gnanadesikan et al. (1995) [14] commented that "One of the thorniest aspects of cluster analysis continue to be the weighting and selection of variables". More recently, Raftery and Dean (2006) [41] pointed out that "Less work has been done on variable selection for clustering than for classification (also called discrimination or supervised learning), perhaps reflecting the fact that the former is a harder problem. In particular, variable selection and dimension reduction in the context of model-based clustering have not received much attention". For variable selection in model-based clustering, most of the recent researches fall into two lines: one is Bayesian (Liu et al. 2003 [30]; Hoff 2006 [18]; Tadesse et al. 2005 [43]; Kim et al. 2006 [25]), while the other is penalized likelihood (Pan and Shen 2007 [39]; Xie et al. 2008 [52]; Wang and Zhu 2008 [50]). The basic statistical models of these approaches are all the same: informative variables are assumed to come from a mixture of Normals, corresponding to clusters, while noise variables coming from a single Normal distribution; they differ in how they are implemented. In particular, the Bayesian approaches are more flexible than the penalized methods (because the latter all require a common diagonal covariance matrix, though our main goal here is to relax this assumption), but they are also computationally more demanding because of their use of MCMC for stochastic search; furthermore, penalized methods enjoy the flexibility of the use of penalty functions, such as to accommodate grouped parameters or variables as to be discussed later. Other recent efforts include the following: Raftery and Dean (2006) [41] considered a sequential, stepwise approach to variable selection in model-based clustering;

however, as acknowledged by the authors, "when the number of variables is vast (e.g., in microarray data analysis when thousands of genes may be the variables being used), the method is too slow to be practical as it stands". Friedman and Meulman (2004) [11] dealt with a more general problem: selecting possibly different subsets of variables and their associated weights for different clusters for non-model-based clustering; Hoff (2004) [17] pointed out that the method might only "pick up the change in variance but not the mean", and advocated the use of his model-based approach (Hoff 2006 [18]). Mangasarian and Wild (2004) [32] proposed the use of $L_1$ penalty for K-median clustering; the idea with the use of $L_1$ penalty is similar to ours, but we consider a more general case with cluster-specific variance parameters.

The penalized methods proposed so far for simultaneous variable selection and model fitting in model-based clustering all assume a common diagonal covariance matrix. For high-dimensional data, it may be necessary to utilize a diagonal covariance matrix for model-based clustering; even for supervised learning, it has been shown that using a diagonal covariance matrix in naive Bayes discriminant analysis or its variants is more effective than that of a more general covariance matrix (Bickel and Levina 2004 [5]; Dudoit et al. 2002 [7]; Tibshirani et al. 2003 [47]). Hence we will restrict our discussion to a diagonal covariance matrix in what follows. On the other hand, the *common* (diagonal) covariance matrix assumption implies that the clusters all have the same size, as in the K-means method (which further assumes that all the clusters are sphere-shaped with a scaled identity matrix as the covariance). Of course, this assumption may be violated in practice. A general argument is the following: it is well known that the variance of gene expression levels is in general a function of the mean expression levels, suggesting possibly varying variances of a gene's expression levels across clusters with different mean expression levels; this point is going to be verified for our real data example. Here we extend the method to allow for cluster-dependent (diagonal) covariance matrices, which is nontrivial and requires a suitable construction of penalty function.

In some applications, there is prior knowledge about grouping variables: some variables function as a group; either all of them or none of them is informative. Yuan and Lin (2006) [54] discussed this issue in the context of penalized regression; they demonstrated convincingly the efficiency gain from incorporating such prior knowledge. On the other hand, in genomic studies of clustering samples through gene expression profiles, it is known that genes function in groups as in biological pathways. Hence, rather than treating genes individually, it seems natural to apply biological knowledge on gene functions to group the genes accordingly in clustering microarray samples, which has not been considered in previous applications of model-based clustering of expression profiles (e.g. Ghosh and Chinnaiyan 2002 [13]; Li and Hong 2001 [27]; McLachlan et al. 2002 [33]; Yeung et al. 2001 [53]). Note that, a few existing works clustered genes by incorporating gene function annotations in a weaker form that did not require either all or none of a group of genes to appear in a final model: Pan (2006b) [38] treated the genes within the same functional group as sharing the same prior probability of being in a cluster, while genes from different groups might not have the same prior probability, in model-based clustering of genes; others took account of gene groups in the definition of a distance metric in other clustering methods (Huang and Pan 2006 [20]). In addition, the aforementioned clustering methods did not allow for variable selection directly, while it is our main aim to consider variable selection, possibly assisted with biological knowledge. This is in line with the currently increasing interest in incorporating biological information on gene functional groups into analysis of detecting differential gene expression (e.g. Pan 2006 [37]; Efron and Tibshirani 2007 [8]; Newton et al. 2007 [36]).

The rest of this article is organized as follows. Section 2 briefly reviews the penalized model-based clustering method with a common diagonal covariance, followed by our proposed methods that allow for cluster-specific diagonal covariance matrices and for grouped variables. The EM algorithms for implementing the proposed methods are also detailed, in which the M-steps characterize the penalized mean and variance estimators with clear effects of shrinkage and thresholding. Simulation results in section 3 and an application to real microarray data in section 4 illustrate the utility of the new methods and their advantages over other methods. Section 5 presents a summary and a short discussion on future work.

## 2. Methods

### 2.1. Mixture model and its penalized likelihood

We have $K$-dimensional observations $x_j, j = 1, \ldots, n$. It is assumed that the data have been standardized to have sample mean 0 and sample variance 1 across the $n$ observations for each variable. The observations are assumed to be (marginally) iid from a mixture distribution with $g$ components: $\sum_{i=1}^{g} \pi_i f_i(x_j; \theta_i)$, where $\theta_i$ is an unknown parameter vector of the distribution for component $i$ while $\pi_i$ is a prior probability for component $i$. To obtain the maximum penalized likelihood estimator (MPLE), we maximize the penalized log-likelihood

$$\log L_P(\Theta) = \sum_{j=1}^{n} \log \left[ \sum_{i=1}^{g} \pi_i f_i(x_j; \theta_i) \right] - p_\lambda(\Theta)$$

where $\Theta$ represents all unknown parameters and $p_\lambda(\Theta)$ is a penalty with regularization parameter $\lambda$. The specific form of $p_\lambda(\Theta)$ depends on the aim of analysis. For variable selection, the $L_1$ penalty is adopted as in the Lasso (Tibshirani 1996 [46]).

Denote by $z_{ij}$ the indicator of whether $x_j$ is from component $i$; that is, $z_{ij} = 1$ if $x_j$ is indeed from component $i$, and $z_{ij} = 0$ otherwise. Because we do not observe which component an observation comes from, $z_{ij}$'s are regarded as missing data. If $z_{ij}$'s could be observed, then the log-penalized-likelihood for complete data is:

$$\log L_{c,P}(\Theta) = \sum_i \sum_j z_{ij}[\log \pi_i + \log f_i(x_j; \theta_i)] - p_\lambda(\Theta). \tag{1}$$

Let $X = \{x_j : j = 1, \ldots, n\}$ represent the observed data. To maximize $\log L_P$, the EM algorithm is often used (Dempster et al. 1977 [6]). The E-step of the EM calculates

$$Q_P(\Theta; \Theta^{(r)}) = E_{\Theta^{(r)}}(\log L_{c,P}|X) = \sum_i \sum_j \tau_{ij}^{(r)} [\log \pi_i + \log f_i(x_j; \theta_i)] - p_\lambda(\Theta), \tag{2}$$

while the M-step maximizes $Q_P$ to update estimated $\Theta$. In the sequel, because $\tau_{ij}$'s always depend on $r$, for simplicity we may suppress the explicit dependence from the notation.

## 2.2. Penalized clustering with a common covariance matrix

Pan and Shen (2007) [39] specified each component $f_i$ as a Normal distribution with a common diagonal covariance structure $V$:

$$f_i(x; \theta_i) = \frac{1}{(2\pi)^{K/2}|V|^{1/2}} \exp\left(-\frac{1}{2}(x - \mu_i)'V^{-1}(x - \mu_i)\right)$$

where $V = diag(\sigma_1^2, \sigma_2^2, \ldots, \sigma_K^2)$, and $|V| = \prod_{k=1}^{K} \sigma_k^2$. They proposed a penalty function $p_\lambda(\Theta)$ with an $L_1$ norm involving the mean parameters alone:

$$p_\lambda(\Theta) = \lambda_1 \sum_{i=1}^{g} \sum_{k=1}^{K} |\mu_{ik}|, \tag{3}$$

where $\mu_{ik}$'s are the components of $\mu_i$, the mean of cluster $i$. Note that, because the data have been standardized to have sample mean 0 and variance 1 for each variable $k$, if $\mu_{1k} = \cdots = \mu_{gk} = 0$, then variable $k$ is noninformative in terms of clustering and can be considered as a noise variable and excluded from the clustering analysis. The $L_1$ penalty function used in (3) can effectively shrink a small $\mu_{ik}$ to be exactly 0.

For completeness and to compare with the proposed methods, we list the EM updates to maximize the above penalized likelihood (Pan and Shen 2007 [39]). We use a generic notation $\Theta^{(r)}$ to represent the parameter estimate at iteration $r$. For the posterior probability of $x_j$'s coming from component $i$, we have

$$\hat{\tau}_{ij}^{(r)} = \frac{\hat{\pi}_i^{(r)} f_i(x_j; \hat{\theta}_i^{(r)})}{f(x_j; \hat{\Theta}^{(r)})} = \frac{\hat{\pi}_i^{(r)} f_i(x_j; \hat{\theta}_i^{(r)})}{\sum_{i=1}^{g} \hat{\pi}_i^{(r)} f_i(x_j; \hat{\theta}_i^{(r)})}, \tag{4}$$

for the prior probability of an observation from the $i^{th}$ component $f_i$,

$$\hat{\pi}_i^{(r+1)} = \sum_{j=1}^{n} \hat{\tau}_{ij}^{(r)}/n, \tag{5}$$

for the variance for variable $k$,

$$\hat{\sigma}_k^{2,(r+1)} = \sum_{k=1}^{K} \sum_{j=1}^{n} \hat{\tau}_{ij}^{(r)}(x_{jk} - \hat{\mu}_{ik}^{(r)})^2/n, \tag{6}$$

and for the mean for variable $k$ in cluster $i$,

$$\hat{\mu}_{ik}^{(r+1)} = \frac{\sum_{j=1}^{n} \hat{\tau}_{ij}^{(r)} x_{jk}}{\sum_{j=1}^{n} \hat{\tau}_{ij}^{(r)}} \left(1 - \frac{\lambda_1 \hat{\sigma}_k^{2,(r)}}{|\sum_{j=1}^{n} \hat{\tau}_{ij}^{(r)} x_{jk}|}\right)_+, \tag{7}$$

with $i = 1, 2, \ldots, g$ and $k = 1, 2, \ldots, K$. Evidently, we have $\hat{\mu}_{ik} = 0$ if $\lambda_1$ is large enough. As discussed earlier, if $\hat{\mu}_{1k} = \hat{\mu}_{2k} = \cdots = \hat{\mu}_{gk} = 0$ for variable $k$, variable $k$ is a noise variable that does not contribute to clustering.

## 2.3. Penalized clustering with unequal covariance matrices

To allow varying cluster sizes, we consider a more general model with cluster-dependent diagonal covariance matrices:

$$f_i(x; \theta_i) = \frac{1}{(2\pi)^{K/2} |V_i|^{1/2}} \exp\left(-\frac{1}{2}(x - \mu_i)' V_i^{-1} (x - \mu_i)\right) \tag{8}$$

where $V_i = diag(\sigma_{i1}^2, \sigma_{i2}^2, \ldots, \sigma_{iK}^2)$, and $|V_i| = \prod_{k=1}^{K} \sigma_{ik}^2$.

As discussed earlier, to realize variable selection, we require that a noise variable have a common mean and a common variance across clusters. Hence, the penalty has to involve both the mean and variance parameters. We shall penalize the mean parameters in the same way as before, while the variance parameters can be regularized in two ways: shrink $\sigma_{ik}^2$ towards 1, or shrink $\log \sigma_{ik}^2$ towards 0.

### 2.3.1. Regularization of variance parameters: scheme one

First, we will use the following penalty for both mean and variance parameters:

$$p_\lambda(\Theta) = \lambda_1 \sum_{i=1}^{g} \sum_{k=1}^{K} |\mu_{ik}| + \lambda_2 \sum_{i=1}^{g} \sum_{k=1}^{K} |\sigma_{ik}^2 - 1|. \tag{9}$$

Again the $L_1$ norm is used to coerce a small estimate of $\mu_{ik}$ to be exactly 0, while forcing an estimate of $\sigma_{ik}^2$ that is close to 1 to be exactly 1. Therefore, if a variable has common mean 0 and common variance 1 across clusters, this variable is effectively treated as a noise variable; this aspect is evidenced from (4), where a noise variable does not contribute to the posterior probability and thus clustering.

Note that penalty (9) differs from the so-called double penalization in non-parametric mixed-effect models for longitudinal and other correlated data (Lin and Zhang 1999 [29]; Gu and Ma 2005 [16]): aside from the obvious differences in the choice of the $L_1$-norm here versus the $L_2$-norm there and in clustering here versus regression there, they penalized fixed- and random-effect parameters, both mean parameters, whereas we regularize variance parameters in addition to mean parameters. Ma et al. (2006) [31] applied such a mixed-effect model to cluster genes with time course (and thus correlated) expression profiles; in addition to the aforementioned differences, a key difference is that their use of penalization was for parameter estimation, not for variable selection as aimed here.

An EM algorithm is derived as follows. The E-step gives $Q_P$ as shown in (2). The M-step maximizes $Q_P$ with respect to the unknown parameters, resulting in the same updating formulas for $\tau_{ij}$ and $\pi_i$ as given in (4) and (5). In Appendix B, we derive the following theorem:

**Theorem 1.** *The sufficient and necessary conditions for $\hat{\mu}_{ik}$ to be a (global) maximizer of $Q_P$ are*

$$\frac{\sum_{j=1}^n \tau_{ij} x_{jk}}{\sum_{j=1}^n \tau_{ij}} = \left(\frac{\lambda_1 \sigma_{ik}^2}{\sum_{j=1}^n \tau_{ij}} + |\hat{\mu}_{ik}|\right) sign(\hat{\mu}_{ik}), \quad \text{if and only if } \hat{\mu}_{ik} \neq 0 \quad (10)$$

*and*

$$\left|\sum_{j=1}^n \tau_{ij} x_{jk}\right|/\sigma_{ik}^2 \leq \lambda_1, \quad \text{if and only if } \hat{\mu}_{ik} = 0, \quad (11)$$

*resulting in a slightly changed formula for the mean parameters*

$$\hat{\mu}_{ik}^{(r+1)} = \frac{\sum_{j=1}^n \hat{\tau}_{ij}^{(r)} x_{jk}}{\sum_{j=1}^n \hat{\tau}_{ij}^{(r)}} \left(1 - \frac{\lambda_1 \hat{\sigma}_{ik}^{2,(r)}}{|\sum_{j=1}^n \hat{\tau}_{ij}^{(r)} x_{jk}|}\right)_+. \quad (12)$$

For the variance parameters, some algebra yields the following theorem:

**Theorem 2.** *The necessary conditions for $\hat{\sigma}_{ik}^2$ to be a local maximizer of $Q_P$ are*

$$\sum_{j=1}^n \tau_{ij} \left(-\frac{1}{2\hat{\sigma}_{ik}^2} + \frac{(x_{jk} - \hat{\mu}_{ik})^2}{2\hat{\sigma}_{ik}^4}\right) = \lambda_2 sign(\hat{\sigma}_{ik}^2 - 1), \quad \text{if } \hat{\sigma}_{ik}^2 \neq 1 \quad (13)$$

*and*

$$\left|\sum_{j=1}^n \tau_{ij} \left(-\frac{1}{2} + \frac{(x_{jk} - \hat{\mu}_{ik})^2}{2}\right)\right| \leq \lambda_2, \quad \text{if } \hat{\sigma}_{ik}^2 = 1. \quad (14)$$

Although a sufficient condition for $\hat{\sigma}_{ik}^2 = 1$ can be derived as a special case of Theorem 5, we do not have any sufficient condition for $\hat{\sigma}_{ik}^2 \neq 1$. Hence, we do not have a simple formula to update $\hat{\sigma}_{ik}^2$. Below we characterize the solution $\hat{\sigma}_{ik}^2$, suggesting a computational algorithm as well as illustrating the effects of shrinkage.

Let $a_{ik} = \lambda_2 \text{sign}(\hat{\sigma}_{ik}^2 - 1)$, $b_i = \sum_{j=1}^n \tau_{ij}/2$, and $c_{ik} = \sum_{j=1}^n \tau_{ij}(x_{jk} - \hat{\mu}_{ik})^2/2$, then (13) reduces to $a_{ik}\sigma_{ik}^4 + b_i \sigma_{ik}^2 - c_{ik} = 0$, while (14) becomes $|b_i - c_{ik}| \leq \lambda_2$. Note that $\tilde{\sigma}_{ik}^2 = c_{ik}/b_i$ is the usual MLE when $\lambda_2 = 0$. It is easy to verify that if $\tilde{\sigma}_{ik}^2 = 1$, then $\hat{\sigma}_{ik}^2 = 1$. Below we consider cases with $\lambda_2 > 0$ and $\tilde{\sigma}_{ik}^2 \neq 1$. It is shown in Appendix B that

1. if $|b_i - c_{ik}| > \lambda_2$,

$$\hat{\sigma}_{ik}^2 = \frac{\tilde{\sigma}_{ik}^2}{\frac{1}{2} + \sqrt{\frac{1}{4} + \frac{\text{sign}(c_{ik}-b_i)\lambda_2 c_{ik}}{b_i^2}}} \quad (15)$$

is the unique maximizer of $Q_P$ and is between 1 and $\tilde{\sigma}_{ik}^2$;

2. if $|b_i - c_{ik}| \leq \lambda_2$,
   (a) if $\tilde{\sigma}_{ik}^2 > 1$, then $\hat{\sigma}_{ik}^2 = 1$ is the unique maximizer;
   (b) if $\tilde{\sigma}_{ik}^2 < 1$, i) if $b_i - c_{ik} < \lambda_2$, then $\hat{\sigma}_{ik}^2 = 1$ is a local maximizer; there may exist another local maximizer between $\tilde{\sigma}_{ik}^2$ and 1; between the two, the one maximizing $Q_P$ is chosen; ii) if $b_i - c_{ik} = \lambda_2$, then either $\hat{\sigma}_{ik}^2 = c_{ik}/\lambda_2 \in (\tilde{\sigma}_{ik}^2, 1)$ (if $\tilde{\sigma}_{ik}^2 < 1/2$) or $\hat{\sigma}_{ik}^2 = 1$ (if $\tilde{\sigma}_{ik}^2 \geq 1/2$) is the unique maximizer.

Naturally the above formulas suggest an updating algorithm for $\sigma_{ik}^2$. Clearly, $\hat{\sigma}_{ik}^2$ has been shrunk towards 1, and can be exactly 1 if, for example, $\lambda_2$ is sufficiently large.

*2.3.2. Regularization of variance parameters: scheme two*

The following penalty is adopted for both mean and variance parameters:

$$p_\lambda(\Theta) = \lambda_1 \sum_{i=1}^{g} \sum_{k=1}^{K} |\mu_{ik}| + \lambda_2 \sum_{i=1}^{g} \sum_{k=1}^{K} |\log \sigma_{ik}^2|. \quad (16)$$

Note that the only difference between (9) and (16) is the penalty of the variance parameters, where $|\sigma_{ik}^2 - 1|$ is replaced by $|\log \sigma_{ik}^2|$, which is used to shrink $\log \sigma_{ik}^2$ to 0 (i.e. $\sigma_{ik}^2$ to 1) if $\log \sigma_{ik}^2$ is close to 0. Therefore, variable selection can be realized as before.

An EM algorithm for the variance parameters is derived as follows.

**Theorem 3.** *The sufficient and necessary conditions for $\hat{\sigma}_{ik}^2$ to be a local maximizer of $Q_P$ are*

$$\sum_{j=1}^{n} \tau_{ij} \left( -\frac{1}{2\hat{\sigma}_{ik}^2} + \frac{(x_{jk} - \hat{\mu}_{ik})^2}{2\hat{\sigma}_{ik}^4} \right) = \frac{\lambda_2 sign(\log \hat{\sigma}_{ik}^2)}{\hat{\sigma}_{ik}^2}, \quad \text{if } \hat{\sigma}_{ik}^2 \neq 1 \quad (17)$$

and

$$\left| \sum_{j=1}^{n} \tau_{ij} \left( -\frac{1}{2} + \frac{(x_{jk} - \hat{\mu}_{ik})^2}{2} \right) \right| \leq \lambda_2, \quad \text{if } \hat{\sigma}_{ik}^2 = 1. \quad (18)$$

If we denote $b_i = \sum_{j=1}^{n} \tau_{ij}/2$ and $c_{ik} = \sum_{j=1}^{n} \tau_{ij}(x_{jk} - \hat{\mu}_{ik})^2/2$, then (17) reduces to $\tilde{\sigma}_{ik}^2 = \hat{\sigma}_{ik}^2(1 + \lambda_2 \text{sign}(\log \hat{\sigma}_{ik}^2)/b_i)$, while (18) becomes $|b_i - c_{ik}| \leq \lambda_2$, where $\tilde{\sigma}_{ik}^2 = c_{ik}/b_i$ is the usual MLE for $\lambda_2 = 0$. Derivations in Appendix B imply that $\text{sign}(\log \hat{\sigma}_{ik}^2) = \text{sign}(\log \tilde{\sigma}_{ik}^2) = \text{sign}(c_{ik} - b_i)$. Combining the two cases, we obtain

$$\hat{\sigma}_{ik}^2 = \left[\frac{\tilde{\sigma}_{ik}^2}{1 + \lambda_2 \text{sign}(c_{ik} - b_i)/b_i} - 1\right] \text{sign}(|b_i - c_{ik}| - \lambda_2)_+ + 1 \qquad (19)$$

The above formula suggests an updating algorithm for $\sigma_{ik}^2$. When $\lambda_2$ is small, $\text{sign}(|b_i - c_{ik}| - \lambda_2)_+ = 1$, $\hat{\sigma}_{ik}^2$ has been shrunk from $\tilde{\sigma}_{ik}^2$ towards 1; when $\lambda_2$ is sufficiently large, $\text{sign}(|b_i - c_{ik}| - \lambda_2)_+ = 0$, then $\hat{\sigma}_{ik}^2$ is exactly 1.

### 2.4. Using adaptive penalization to reduce bias

We can adopt the idea of adaptive penalization, as proposed by Zou (2006) [57] for regression, in the present context. Following Pan et al. (2006) [40], we use a weighted $L_1$ penalty function

$$p_\lambda(\Theta) = \lambda_1 \sum_{i=1}^{g} \sum_{k=1}^{K} w_{ik} |\mu_{ik}| + \lambda_2 \sum_{i=1}^{g} \sum_{k=1}^{K} v_{ik} |\sigma_{ik}^2 - 1|, \qquad (20)$$

where $w_{ik} = 1/|\hat{\mu}_{ik}|^w$ and $v_{ik} = 1/|\hat{\sigma}_{ik}^2 - 1|^w$ with $w \geq 0$, and $\hat{\mu}_{ik}$ and $\hat{\sigma}_{ik}^2$ are the MPLE obtained in section 2.3.1; we also tried the usual MLE in $w_{ik}$ and $v_{ik}$, but it did not work well in simulations, hence we skip its discussion; we only consider $w = 1$. The EM updates are slightly modified for the purpose: we only need to replace $\lambda_1$ and $\lambda_2$ by $\lambda_1 w_{ik}$ and $\lambda_2 v_{ik}$ respectively, while keeping other updates unchanged.

The main idea of adaptive penalization is to reduce the bias of the MPLE associated with the standard $L_1$ penalty: as can be seen clearly, if an initial estimate $|\hat{\mu}_{ik}|$ is larger, then the resulting estimate is shrunk less towards 0; similarly for the variance parameter.

### 2.5. Penalized clustering with grouped variables

Now we consider a situation where candidate variables can be grouped based on the prior belief that either all the variables in the same group or none of them are informative to clustering. Following the same idea of the grouped Lasso of Yuan and Lin (2006) [54], we construct a penalty for this purpose here.

Suppose that the variables are partitioned into $M$ groups with the corresponding mean parameters $\mu_i = (\mu_{i1}, \mu_{i2}, \ldots, \mu_{iK})\prime = (\mu_i^1\prime, \mu_i^2\prime, \ldots, \mu_i^M\prime)\prime$, $\dim(\mu_i^m) = k_m$, and $\sum_{m=1}^{M} k_m = K$. Accordingly, we decompose $x_j = (x_j^1\prime, x_j^2\prime, \ldots, x_j^M\prime)\prime$ and $V_i = diag(\sigma_{i1}^2, \sigma_{i2}^2, \ldots, \sigma_{iK}^2) = diag(V_{i1}, V_{i2}, \ldots, V_{iM})$ with $V_{im}$ as a $k_m \times k_m$ diagonal matrix, and $\sigma_{i,m}^2$ is the column vector containing the diagonal elements of matrix $V_{im}$.

For grouping mean parameters, we will use a penalty $p_\lambda(\Theta)$ containing

$$\lambda_1 \sum_{i=1}^{g} \sum_{m=1}^{M} \sqrt{k_m} \|\mu_i^m\|$$

for the mean parameters, where $\|\mathbf{v}\|$ is the $L_2$ norm of vector $\mathbf{v}$. Accordingly, we use

$$\lambda_2 \sum_{i=1}^{g} \sum_{m=1}^{M} \sqrt{k_m} \|\sigma_{i,m}^2 - \mathbf{1}\|$$

as a penalty for grouped variance parameters. Note that we do not have to group both means and variances at the same time. For instance, we may group only means but not variances: we will thus use the second term in (9) as the penalty for variance parameters while retaining the above penalty for grouped mean parameters.

The E-step of the EM yields $Q_P$ with the same form as (2). Next we derive the updating formulas for the mean and variance parameters in the M-step.

### 2.5.1. Grouping mean parameters

If the penalty for grouped means is used, we have the following result.

**Theorem 4.** *The sufficient and necessary conditions for $\mu = (\mu_i^m), i = 1, 2, \ldots, g$ and $m = 1, 2, \ldots, M$ to be a unique maximizer of $Q_P$ are*

$$V_{im}^{-1} \left( \sum_{j=1}^{n} \tau_{ij} x_j^m - \left( \sum_{j=1}^{n} \tau_{ij} \right) \mu_i^m \right) = \lambda_1 \sqrt{k_m} \frac{\mu_i^m}{\|\mu_i^m\|}, \quad \text{if and only if } \mu_i^m \neq \mathbf{0}, \tag{21}$$

*and*

$$\left\| \sum_{j=1}^{n} \tau_{ij} x_j^m \prime V_{im}^{-1} \right\| \leq \lambda_1 \sqrt{k_m}, \quad \text{if and only if } \mu_i^m = \mathbf{0}, \tag{22}$$

*yielding*

$$\hat{\mu}_i^m = \left( sign \left( 1 - \frac{\lambda_1 \sqrt{k_m}}{\| \sum_{j=1}^{n} \tau_{ij} x_j^m \prime V_{im}^{-1} \|} \right) \right)_+ \nu_i^m \tilde{\mu}_i^m \tag{23}$$

*where $\nu_i^m = \left( \mathbf{I} + \frac{\lambda_1 \sqrt{k_m}}{\sum_{j=1}^{n} \tau_{ij} \|\hat{\mu}_i^m\|} V_{im} \right)^{-1}$, and $\tilde{\mu}_i^m = \sum_{j=1}^{n} \tau_{ij} x_j^m / \sum_{j=1}^{n} \tau_{ij}$ is the usual MLE.*

It is clear that, due to thresholding, $\hat{\mu}_i^m = \mathbf{0}$ when, for example, $\lambda_1$ is sufficiently large. Noting that $\nu_i^m$ depends on $\hat{\mu}_i^m$, we use (23) iteratively to update $\hat{\mu}_i^m$.

### 2.5.2. Grouping variance parameters

If the penalty for grouped variances is used, we have the following theorem:

**Theorem 5.** *The sufficient and necessary conditions for $\sigma_{i,m}^2 = \mathbf{1}$, $i = 1, 2, \ldots, g$ and $m = 1, 2, \ldots, M$, to be a local maximizer of $Q_P$ are*

$$\begin{cases} \left\| \sum_{j=1}^n \tau_{ij} \left[ \frac{1}{2}\mathbf{1} - \frac{1}{2}(x_j^m - \mu_i^m)^2 \right] \right\| \leq \lambda_2 \sqrt{k_m} & \text{if } \sum_{j=1}^n \tau_{ij} \left[ \frac{1}{2}\mathbf{1} - (x_j^m - \mu_i^m)^2 \right] \leq \mathbf{0}; \\ \left\| \sum_{j=1}^n \tau_{ij} \left[ \frac{1}{2}\mathbf{1} - \frac{1}{2}(x_j^m - \mu_i^m)^2 \right] \right\| < \lambda_2 \sqrt{k_m} & \text{otherwise.} \end{cases}$$

(24)

*The necessary condition for $\sigma_{i,m}^2 \neq \mathbf{1}$ to be a local maximizer of $Q_P$ is*

$$\sum_{j=1}^n \tau_{ij} \left[ -\frac{1}{2\sigma_{i,m}^2} + \frac{1}{2(\sigma_{i,m}^2)^2}(x_j^m - \mu_i^m)^2 \right] = \frac{\lambda_2 \sqrt{k_m}(\sigma_{i,m}^2 - \mathbf{1})}{\|\sigma_{i,m}^2 - \mathbf{1}\|}. \quad (25)$$

It is clear that $\sigma_{i,m}^2 = \mathbf{1}$ when, for example, $\lambda_2$ is large enough. It is also easy to verify that (24) and (25) reduce to the same ones for non-grouped variables when $k_m = 1$. To solve (24) and (25), we develop the following algorithm. Let $a_{im} = \lambda_2 \sqrt{k_m}/\|\sigma_{i,m}^2 - \mathbf{1}\|$, $\mathbf{b_i} = (\sum_{j=1}^n \tau_{ij}/2)\mathbf{1}$ and $\mathbf{c_{im}} = \sum_{j=1}^n \tau_{ij}(x_j^m - \mu_i^m)^2/2$. Consider any $k'$th component $\sigma_{ik'}^2$ of $\sigma_{i,m}^2$; correspondingly, $b_{ik'}$ and $c_{imk'}$ are the $k'$th components of $\mathbf{b_i}$ and $\mathbf{c_{im}}$, respectively. In Appendix B, treating $a_{im}$ as a constant (i.e. by plugging-in a current estimate of $\sigma_{i,m}^2$), we show the following cases. i) If $\tilde{\sigma}_{ik'}^2 = 1$, then $\hat{\sigma}_{ik'}^2 = 1$ is a maximizer of $Q_P$ as other $\sigma_{ik}^2$'s for $\forall k \neq k'$ are fixed. ii) If $\tilde{\sigma}_{ik'}^2 = c_{imk'}/b_{ik'} > 1$, there exists only one real root satisfying $\hat{\sigma}_{ik'}^2 \in (1, \tilde{\sigma}_{ik'}^2)$; a bisection search can be used to find the root. iii) If $\tilde{\sigma}_{ik'}^2 = c_{imk'}/b_{ik'} < 1$, the real roots must be inside $(\tilde{\sigma}_{ik'}^2, 1)$, hence a bisection search can be used to find a root; once a root is obtained, the other two real roots, if exist, can be obtained through a closed-form expression; we choose the real root that maximizes $Q_P$ (while other $\sigma_{ik}^2$ for $k \neq k'$ are fixed at their current estimates) as the new estimate of $\sigma_{ik'}^2$. After cycling through all $k'$, we update $a_{im}$ with the new estimate. Then the above process is iterated.

### 2.5.3. Other grouping schemes

To save space, we briefly discuss grouping variables under a common diagonal covariance matrix, for which only mean parameters need to be regularized. The EM updating formula for the mean parameters remains the same as in (23) except that the cluster-specific covariance $V_{im}$ there is replaced by a common $V_m$; updating formulas for the other parameters remain unchanged. Simulation results (see Xie et al. (2008) [52]) demonstrated its improved performance over its counterpart without grouping. In addition, we can also group the mean parameters for the same variable (or gene) across clusters (Wang and Zhu 2008 [50]), and combine it with grouping variables (Xie et al. 2008 [52]).

The grouping scheme discussed so far follows the grouped Lasso of Yuan and Lin (2006) [54], which is a special case of the Composite Absolute Penalties (CAP) of Zhao et al. (2006) [56]. In Appendix A, we derive the results with the CAP, including using both schemes on regularizing the variance parameters.

### 2.6. Model selection

To introduce penalization, following Pan and Shen (2007) [39] and Pan et al. (2006) [40], we propose a modified BIC as the model selection criterion:

$$BIC = -2\log L(\hat{\Theta}) + \log(n)d_e$$

where $d_e = g + K + gK - 1 - q$ is the effective number of parameters with $q = \#\{(i,k) : \mu_{ik} = 0, \sigma_{ik}^2 = 1\}$. The definition of $d_e$ follows from that in $L_1$-penalized regression (Efron et al. 2004 [9]; Zou et al. 2004 [59]). This modified BIC is used to select the number of clusters $g$ and the penalization parameters $(\lambda_1, \lambda_2)$ jointly. We propose using a grid search to estimate the optimal $(g, \hat{\lambda}_1, \hat{\lambda}_2)$ as the one with the minimum BIC.

For any given $(g, \lambda_1, \lambda_2)$, because of possibly many local maxima for the mixture model, we run an EM algorithm multiple times with random starts. For our numerical examples, we randomly started the K-means and used the K-means' results as an initialization for the EM. From the multiple runs, we selected the one giving the maximal penalized log-likelihood as the final result for the given $(g, \lambda_1, \lambda_2)$.

## 3. Simulations

### 3.1. A common covariance versus unequal covariances

#### 3.1.1. Case I

We first considered four simple set-ups: the first was a null case with $g = 1$; for the other three, $g = 2$, corresponding to only mean, only variance, and both mean and variance differences between the two clusters. Specifically, we generated 100 simulated datasets for each set-up. In each dataset, there were $n = 100$ observations, each of which contained $K = 300$ variables. Set-up 1) is a null case: all the variables had a standard normal distribution $N(0, 1)$, thus there was only a single cluster. For each of the other three set-ups, there were two clusters. One cluster contained 80 observations while the other contained 20; while 279 variables were noises distributed as $N(0, 1)$, the other 21 variables were informative: each of the 21 variables were distributed as 2) $N(0, 1)$ in cluster 1 versus $N(1.5, 1)$ in cluster 2; 3) $N(0, 1)$ versus $N(0, 2)$; 4) $N(0, 1)$ versus $N(1.5, 2)$ for the three set-ups respectively.

For each simulated dataset, we fitted a series of models with the three numbers of components $g = 1, 2, 3$ and various values of penalization parameter(s). For comparison, we considered both the equal covariance and unequal covariance mixture models (8); for the former, we considered the unpenalized method ($\lambda_1 = 0$) corresponding to no variable selection and penalized method using BIC to select $\lambda$; similarly, for the latter we considered five cases corresponding to fixing or selecting one or two of $(\lambda_1, \lambda_2)$: no variable selection with $(\lambda_1, \lambda_2) = (0, 0)$, penalizing only mean parameters with $(\lambda_1, \lambda_2) = (\hat{\lambda}_1, 0)$, penalizing only variance parameters with $(\lambda_1, \lambda_2) = (0, \hat{\lambda}_2)$, and our proposed two methods of penalizing both mean and variance parameters with $(\lambda_1, \lambda_2) = (\hat{\lambda}_1, \hat{\lambda}_2)$. We also compared the numbers of predicted noise variables among the true 21 informative ($z_1$) and 279 noise variables ($z_2$).

TABLE 1
*Simulation case I: frequencies of the selected numbers (g) of clusters, and mean numbers of predicted noise variables among the true informative ($z_1$) and noise variables ($z_2$). Here $N_1$ and $N_2$ were the frequencies of selecting UnequalCov($\hat{\lambda}_1,\hat{\lambda}_2$) (with variance regularization scheme one) and EqualCov($\hat{\lambda}_1$) by BIC, respectively. UnequalCov($\hat{\lambda}_1,\hat{\lambda}_2$) (logvar) used variance regularization scheme two. For set-up 1, the truth was $g = 1$, $z_1 = 21$ and $z_2 = 279$; for others, $g = 2$, $z_1 = 0$ and $z_2 = 279$*

| | | UnequalCov($\lambda_1,\lambda_2$) | | | | | | | | | EqualCov($\lambda_1$) | | | | BIC | |
|---|---|---|---|---|---|---|---|---|---|---|---|---|---|---|---|---|
| | | (0,0) | ($\hat{\lambda}_1$,0) | (0,$\hat{\lambda}_2$) | ($\hat{\lambda}_1,\hat{\lambda}_2$) | | | ($\hat{\lambda}_1,\hat{\lambda}_2$)(logvar) | | | (0) | ($\hat{\lambda}_1$) | | | | |
| Set-up | g | N | N | N | N | $z_1$ | $z_2$ | N | $z_1$ | $z_2$ | N | N | $z_1$ | $z_2$ | $N_1$ | $N_2$ |
| | 1 | 99 | 83 | 99 | 100 | 21.0 | 279.0 | 100 | 21.0 | 279.0 | 100 | 100 | 21.0 | 279.0 | 0 | 100 |
| 1 | 2 | 0 | 3 | 0 | 0 | - | - | 0 | - | - | 0 | 0 | - | - | 0 | 0 |
| | 3 | 1 | 14 | 1 | 0 | - | - | 0 | - | - | 0 | 0 | - | - | 0 | 0 |
| | 1 | 99 | 89 | 99 | - | - | - | - | - | - | 100 | 0 | - | - | 0 | 0 |
| 2 | 2 | 0 | 1 | 0 | 100 | 0.03 | 276.0 | 100 | 0.03 | 276.0 | 0 | 87 | 0.03 | 275.1 | 0 | 87 |
| | 3 | 1 | 10 | 1 | - | - | - | - | - | - | 0 | 13 | 9.8 | 0.0 | 0 | 13 |
| | 1 | 97 | 74 | 97 | 52 | 21.0 | 279.0 | 43 | 21.0 | 279.0 | 100 | 100 | 21.0 | 279.0 | 48 | 4 |
| 3 | 2 | 0 | 3 | 0 | 42 | 5.4 | 276.8 | 48 | 3.4 | 275.1 | 0 | 0 | - | - | 42 | 0 |
| | 3 | 3 | 23 | 3 | 6 | 6.8 | 277.8 | 9 | 3.6 | 276.3 | 0 | 0 | - | - | 6 | 0 |
| | 1 | 97 | 74 | 97 | 0 | - | - | 0 | - | - | 100 | 4 | 21.0 | 279.0 | 0 | 0 |
| 4 | 2 | 0 | 3 | 0 | 98 | 0.2 | 275.9 | 97 | 0.1 | 275.2 | 0 | 88 | 2.9 | 276.2 | 90 | 2 |
| | 3 | 3 | 23 | 3 | 2 | 0.5 | 276.5 | 3 | 0.3 | 274.0 | 0 | 8 | 6.0 | 276.8 | 8 | 0 |
| | 1 | 100 | 100 | 100 | 52 | 21.0 | 279.0 | 51 | 21.0 | 279.0 | 100 | 100 | 21.0 | 279.0 | 0 | 63 |
| 3 | 2 | 0 | 0 | 0 | 38 | 2.5 | 277.4 | 40 | 2.1 | 275.0 | 0 | 0 | - | - | 34 | 0 |
| (adapt) | 3 | 0 | 0 | 0 | 10 | 3.5 | 277.6 | 9 | 3.4 | 275.4 | 0 | 0 | - | - | 3 | 0 |

The frequencies of selecting $g = 1$ to 3 based on 100 simulated datasets are shown in Table 1. First, our proposed methods performed best, in general, in terms of selecting both the correct number of clusters and relevant variables. For example, it selected fewest noise variables and most informative variables. Second, no variable selection (i.e. no penalization) led to incorrectly selecting $g = 1$ for the three non-null set-ups. Third, penalizing only the mean parameters could not distinguish the two clusters differing only in variance as in set-up 3. Fourth, between the two regularization schemes for the variance parameters, based on both cluster detection and sample assignment (Table 2), the two gave comparable results, though scheme two with log-variance performed slightly better.

The results for adaptive penalty for set-up 3 are detailed in row 3(adapt) of Table 1, which are similar to that of using the $L_1$-norm penalty in terms of both variable and cluster number selection. Since the performance of adaptive penalty might heavily depend on the choice of weights (or initial estimates), we expect improved performance if better weights can be used.

TABLE 2

Sample assignments for $\hat{g} = 2$ and (adjusted) Rand index (RI/aRI) values for the two regularization schemes for the variance parameters for simulation set-ups 2-4. #Corr represents the average number of the samples from a true cluster correctly assigned to an estimated cluster

| | | Sample assignments, $\hat{g} = 2$ | | Rand Index | | | | | | | |
|---|---|---|---|---|---|---|---|---|---|---|---|
| | | Cluster 1 ($n=80$) | Cluster 2 ($n=20$) | $\hat{g}=1$ | | $\hat{g}=2$ | | $\hat{g}=3$ | | Overall | |
| Set-up | Methods | #Corr | #Corr | RI | aRI | RI | aRI | RI | aRI | RI | aRI |
| 2 | $L_1$(var-1) | 78.8 | 19.0 | - | - | 1.00 | 0.99 | - | - | 1.00 | 0.99 |
| | $L_1$(logvar) | 78.8 | 19.0 | - | - | 1.00 | 0.99 | - | - | 1.00 | 0.99 |
| 3 | $L_1$(var-1) | 74.6 | 15.1 | 0.68 | 0.0 | 0.89 | 0.75 | 0.87 | 0.70 | 0.78 | 0.36 |
| | $L_1$(logvar) | 76.9 | 16.6 | 0.68 | 0.0 | 0.94 | 0.85 | 0.95 | 0.88 | 0.83 | 0.49 |
| 4 | $L_1$(var-1) | 78.4 | 19.0 | - | - | 0.99 | 0.98 | 0.97 | 0.93 | 0.99 | 0.98 |
| | $L_1$(logvar) | 78.8 | 19.0 | - | - | 1.00 | 0.99 | 0.99 | 0.98 | 1.00 | 0.99 |

TABLE 3

Simulation case II. The mean numbers of the predicted noise variables as in each of the first three groups of true informative and the other group of noise variables were given by $z_1$-$z_4$; the truth was that $g = 2$, $z_1 = z_2 = z_3 = 0$ and $z_4 = 300 - 3K_1$

| | | UnequalCov($\lambda_1, \lambda_2$) | | | | | | | EqualCov($\lambda_1$) | | | | | |
|---|---|---|---|---|---|---|---|---|---|---|---|---|---|---|
| | | (0,0) | $(\hat{\lambda}_1, 0)$ | $(0, \hat{\lambda}_2)$ | $(\hat{\lambda}_1, \hat{\lambda}_2)$ | | | | (0) | $(\hat{\lambda}_1)$ | | | | |
| $K_1$ | $g$ | $N$ | $N$ | $N$ | $N$ | $z_1$ | $z_2$ | $z_3$ | $z_4$ | $N$ | $N$ | $z_1$ | $z_2$ | $z_3$ | $z_4$ |
| 5 | 1 | 96 | 76 | 96 | 48 | 5.0 | 5.0 | 5.0 | 285.0 | 100 | 74 | 5.0 | 5.0 | 5.0 | 285 |
| | 2 | 0 | 1 | 0 | 42 | 2.0 | 1.8 | 1.5 | 283.5 | 0 | 15 | 0.0 | 4.5 | 0.5 | 279.7 |
| | 3 | 4 | 23 | 4 | 10 | 0.2 | 0.4 | 0.0 | 283.9 | 0 | 11 | 0.0 | 4.2 | 0.3 | 279.5 |
| 7 | 1 | 96 | 81 | 96 | 11 | 7.0 | 7.0 | 7.0 | 279.0 | 100 | 26 | 7.0 | 7.0 | 7.0 | 279 |
| | 2 | 2 | 4 | 2 | 81 | 0.8 | 1.0 | 0.5 | 276.5 | 0 | 54 | 0.0 | 6.3 | 0.7 | 274.7 |
| | 3 | 2 | 15 | 2 | 8 | 0.0 | 0.6 | 0.0 | 277.0 | 0 | 20 | 0.0 | 6.6 | 0.6 | 275.3 |
| 10 | 1 | 99 | 86 | 99 | 0 | - | - | - | - | 100 | 1 | 10.0 | 10.0 | 10.0 | 270.0 |
| | 2 | 0 | 2 | 0 | 94 | 0.2 | 0.9 | 0.1 | 266.3 | 0 | 81 | 0.01 | 9.0 | 0.8 | 266.6 |
| | 3 | 1 | 12 | 1 | 6 | 0.0 | 0.3 | 0.0 | 266.8 | 0 | 18 | 0.0 | 9.1 | 0.7 | 267.6 |

*3.1.2. Case II*

We considered a more practical scenario that combined clusters' differences in means or variances or both for informative variables. As before, for each dataset, $n = 100$ observations were divided into two clusters with 80 and 20 observations respectively; among the $K = 300$ variables, only $3K_1$ were informative while the remaining $K - 3K_1$ were noises; The first, second and third $K_1$ informative variables were distributed as i) $N(0,1)$ for cluster 1 versus $N(1.5,1)$ for cluster 2, ii) $N(0,1)$ versus $N(0,2)$, iii) $N(0,1)$ versus $N(1.5,2)$, respectively; any noise variable was distributed $N(0,1)$. We considered $K_1 = 5, 7$, and 10.

The results are shown in Table 3. Again it is clear that our proposed method worked best: it most frequently selected the correct number of clusters ($g = 2$), and used most informative variables while being able to weed out most noise variables. As expected, using noise variables, as in non-penalized methods without variable selection, tended to mask out the existence of the two clusters.

Table 4
Simulation case II for grouped variables

| $K_1$ | $g$ | Grouping means only | | | | | Means and variances | | | | |
|---|---|---|---|---|---|---|---|---|---|---|---|
| | | $N$ | $z_1$ | $z_2$ | $z_3$ | $z_4$ | $N$ | $z_1$ | $z_2$ | $z_3$ | $z_4$ |
| | 1 | 21 | 5.0 | 5.0 | 5.0 | 285.0 | 14 | 5.0 | 5.0 | 5.0 | 285.0 |
| 5 | 2 | 62 | 0.2 | 0.5 | 0.2 | 283.4 | 71 | 0.0 | 0.0 | 0.0 | 284.6 |
| | 3 | 17 | 0.0 | 0.7 | 0.0 | 283.2 | 15 | 0.0 | 0.0 | 0.0 | 284.7 |
| | 1 | 2 | 7.0 | 7.0 | 7.0 | 279.0 | 1 | 7.0 | 7.0 | 7.0 | 279.0 |
| 7 | 2 | 90 | 0.0 | 0.6 | 0.0 | 277.4 | 92 | 0.0 | 0.0 | 0.0 | 278.9 |
| | 3 | 8 | 0.0 | 0.8 | 0.0 | 277.4 | 7 | 0.0 | 0.0 | 0.0 | 279.0 |
| | 1 | 0 | - | - | - | - | 0 | - | - | - | - |
| 10 | 2 | 96 | 0.0 | 0.7 | 0.0 | 268.3 | 98 | 0.0 | 0.0 | 0.0 | 269.8 |
| | 3 | 4 | 0.0 | 0.8 | 0.0 | 266.8 | 2 | 0.0 | 0.0 | 0.0 | 270.0 |

## 3.2. Grouping variables

We grouped variables for each simulated data under case II. We used correct groupings: the informative variables were grouped together, and so were the noise variables; the group sizes were 5, 7 and 5 for $K_1 = 5, 7$ and 10 respectively. Table 4 displays the results for grouped variables. Compared to Table 1, it is clear that grouping variables improved the performance over non-grouped one in terms of more frequently selecting the correct number $g = 2$ of the clusters as well as better selecting relevant variables. Furthermore, grouping both means and variances performed better than grouping means alone.

## 4. Example

### 4.1. Data

A leukemia gene expression dataset (Golub et al. 1999 [15]) was used to demonstrate the utility of our proposed method and to compare with other methods. The (training) data contained 38 patients, among which 11 were AML (acute myeloid leukemia) while the remaining were ALL (acute lymphoblastic leukemia) samples; ALL samples consisted of two subtypes: 8 T-cell and 19 B-cell samples. For each sample, the expression levels of 7129 genes were measured by an Affymetrix microarray. Following Dudoit et al. (2002) [7], we pre-processed the data in the following steps: 1) truncation: any expression level $x_{jk}$ was truncated below at 1 if $x_{jk} < 1$, and above at 16,000 if $x_{jk} > 16,000$; 2) filtering: any gene was excluded if its $max/min \leq 5$ and $max - min \leq 500$, where $max$ and $min$ were the maximum and minimum expression levels of the gene across all the samples. Finally, as a preliminary gene screening, we selected the top 2000 genes with the largest sample variances across the 38 samples.

Table 5
*Clustering results for Golub's data with the number of components (g) selected by BIC*

| Methods | UnequalCov($\lambda_1, \lambda_2$) | | | | | | | | EqualCov($\lambda_1$) | | | | | | | | | | | |
|---|---|---|---|---|---|---|---|---|---|---|---|---|---|---|---|---|---|---|---|---|
| | (0,0) | | ($\hat{\lambda}_1, \hat{\lambda}_2$) | | | | (0) | | | ($\hat{\lambda}_1$) | | | | | | | | | | |
| RI/aRI | 0.73/0.46 | | 0.85/0.65 | | | | 0.70/0.37 | | | 0.70/0.25 | | | | | | | | | | |
| BIC | 71225 | | 52198 | | | | 75411 | | | 63756 | | | | | | | | | | |
| Clusters | 1 | 2 | 1 | 2 | 3 | 4 | 1 | 2 | 3 | 1 | 2 | 3 | 4 | 5 | 6 | 7 | 8 | 9 | 10 | 11 |
| Samples(#) | | | | | | | | | | | | | | | | | | | | |
| ALL-T(8) | 8 | 0 | 0 | 0 | 8 | 0 | 8 | 0 | 0 | 7 | 0 | 0 | 0 | 0 | 0 | 1 | 0 | 0 | 0 | 0 |
| ALL-B(19) | 17 | 2 | 11 | 1 | 0 | 7 | 0 | 8 | 11 | 0 | 4 | 5 | 2 | 4 | 0 | 0 | 1 | 1 | 1 | 1 |
| AML(11) | 0 | 11 | 0 | 11 | 0 | 0 | 0 | 0 | 11 | 0 | 0 | 0 | 0 | 7 | 4 | 0 | 0 | 0 | 0 | 0 |

## 4.2. No grouping

### 4.2.1. Model-based clustering methods

Table 5 displays the clustering results: the two penalized methods selected 4 and 11 clusters, respectively, while the two standard methods chose 2 and 3 clusters, respectively. For the new penalized method, we show the results for scheme one of regularizing the variance parameters; the other scheme and the adaptive penalization both gave similar results, and hence are skipped. In terms of discriminating between the luekemia subtypes, obviously the new penalized method performed best: only one ALL B-cell sample was mixed into the AML group, while others formed homogeneous groups. In contrast, with a large number of clusters, the $L_1$ method with an equal covariance still misclassified 4 ALL B-cell samples as AML. The two standard methods perhaps under-selected the number of clusters, resulting in 11 and 10 mis-classified samples, respectively. Unsurprisingly, based on the Rand index (Rand 1971 [42]) (or adjusted Rand index (Hubert and Arabie 1985 [22])), the new method was a winner with the largest value at 0.85 (0.65), compared to 0.73 (0.46), 0.70 (0.37) and 0.70 (0.25) of the other three methods. In addition, judged by BIC, the new penalized method also outperformed the other methods with the smallest BIC value of 52198. Finally, the new penalized method used only 1728 genes, while penalizing only means with a common covariance matrix used 1821 genes; the other two standard methods used all 2000 genes.

Figure 1 displays the estimated means and variances of the genes in different clusters. Panels a)-c) clearly show that the genes may have different variance estimates across the clusters, though many of them were shrunk to be exactly to be one, as expected. Note that, due to the standardization of the data yielding an overall sample variance one for each gene, we do not observe any gene with the variance estimates more than one in two or more clusters. Panels d)-f) confirmed that there appears a monotonic relationship between the mean and variance, as well-known in the microarray literature (e.g. Huang and Pan 2002 [19]); the Pearson correlation coefficients were estimated to be 0.64, 0.69, 0.65 and 0.63 for the four clusters respectively. Hence, it lends an indirect support for the use of cluster-specific covariance matrices: if it is accepted that the genes have varying mean parameters across clusters, then their variance parameters are expected to change too.

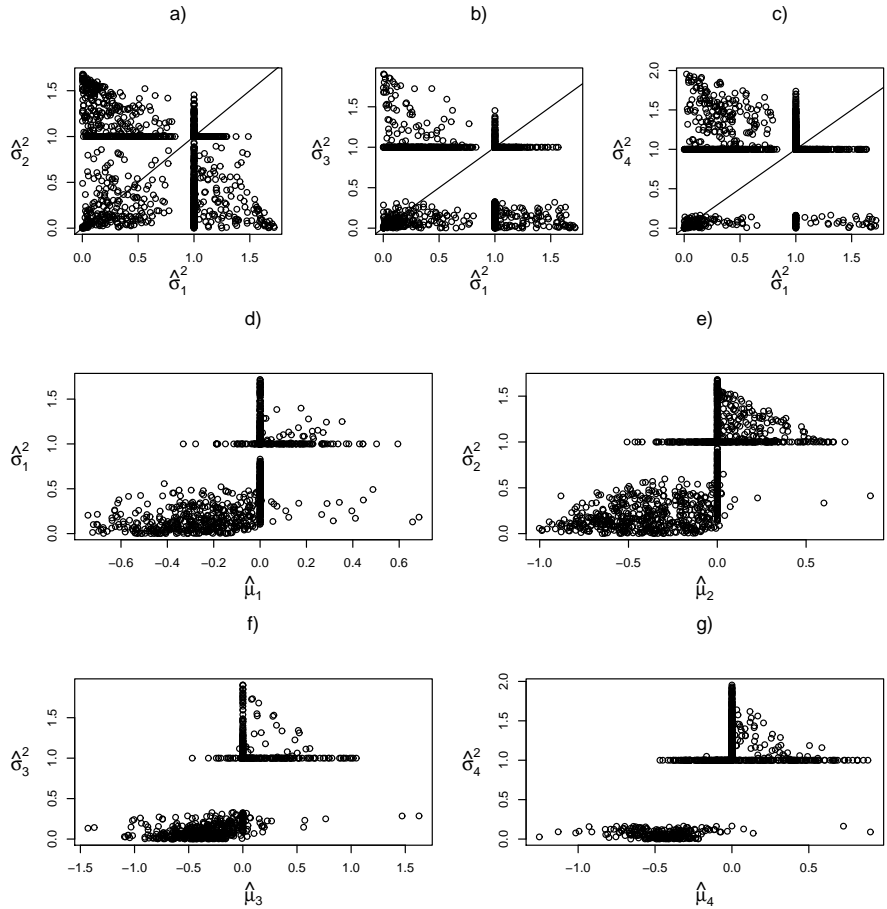

FIG 1. *Scatter plots of the estimated means and variances by the new penalized method. Panels a)–c) are scatter plots of the estimated variances in cluster 1 versus those in cluster 2, 3 and 4, respectively; panels d)–g) are the scatter plots of the estimated means versus estimated variances for the four clusters respectively.*

Next we examine a few genes in more details. CST3 (cystatin c, M23197) and ZYX (zyxin, X95735) were in the top 50 genes ranked by Golub et al. (1999) [15], and two of the 17 genes selected by Antonov et al. (2004) [2] to discriminate between the AML/ALL subtypes. In addition, the two genes, together with MAL (X76223), were also identified among the top 20 genes used in the classifier by Liao et al. (2007) [28] to predict leukemia subtypes. Bardi et al. (2004) [4] used CST3 to assess glomerular function among children with leukemia and solid tumors and found that CST3 was a suitable marker. Koo et al. (2006) [26] reviewed the literature showing the relevance of MAL to T-cell ALL and concluded that it might play an important role in T-cell activations. Baker et al. (2006) [3] and Wang et al. (2005) [49] identified ZYX as the most significant gene for classifying AML/ALL subtypes. Tycko et al. (1991) [48] found that LCK (M26692) was related to activated T cells and thus it might contribute to the formation of human cancer. Wright et al. (1994) [51] studied the mutation of LCK and concluded that it probably played a role in some human T-cell leukemia.

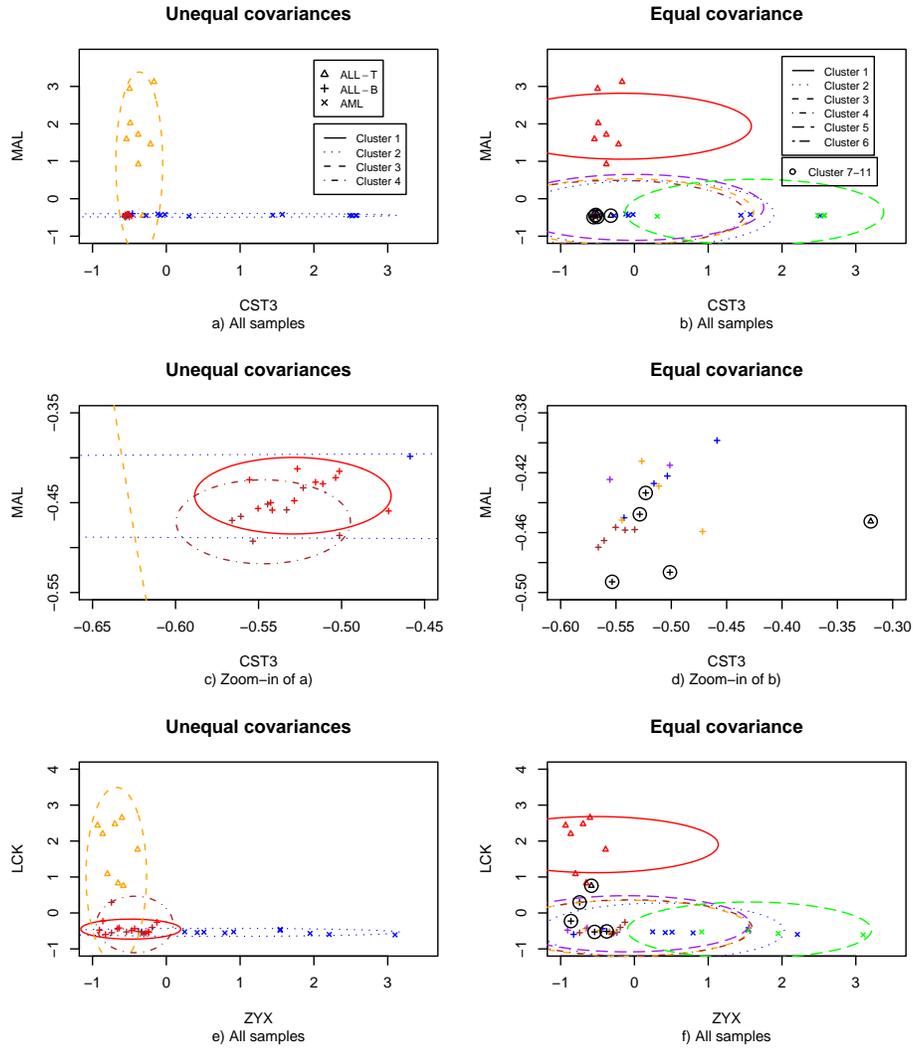

Fig 2. *Observed expression levels of two pairs of genes and the corresponding clusters found by the two penalized methods.*

In Figure 2, panels *c)–d)* are the zoom-in versions of the left bottom of *a)–b)*, the plots of gene pair CST3 and MAL for all samples for the two penalized methods respectively, while $e$)–$f$) are for gene pair LCK and ZYX with all samples. Given two genes, their observed expression levels, along with the 95% confidence region of the center for each cluster, were plotted. The penalized method with an equal covariance found 11 clusters, among which 5 clusters had only one sample, and the remaining 6 clusters had more than 2 samples; for clarity, we only plotted the confidence regions of the centers of the six largest clusters. Panels a) and e) clearly show evidence of varying variances, and thus cluster-specific covariance matrices: for example, MAL was highly expressed with a large dispersion for ALL-T samples, so was CST3 for AML samples, in contrast to the low expression of both genes for ALL-B samples, suggesting varying cluster sizes. It also clearly illustrates why there were so many clusters if an equal covariance model was used: the large number of the equally-sized clusters were used to approximate the three or four size-varying true clusters. Panel c) also suggests an explanation to the use of two clusters for ALL-B samples by the new penalized method: the expression levels of MAL and CST3 were positively correlated, giving a cluster not parallel with either coordinate; on the other hand, use of a diagonal covariance matrix in the penalized method implied a cluster parallel to one of the two coordinates. Hence, two coordinate-parallel clusters were needed to approximate the non-coordinate-parallel true cluster; a non-diagonal covariance matrix might give a more parsimonious model (i.e. with fewer clusters).

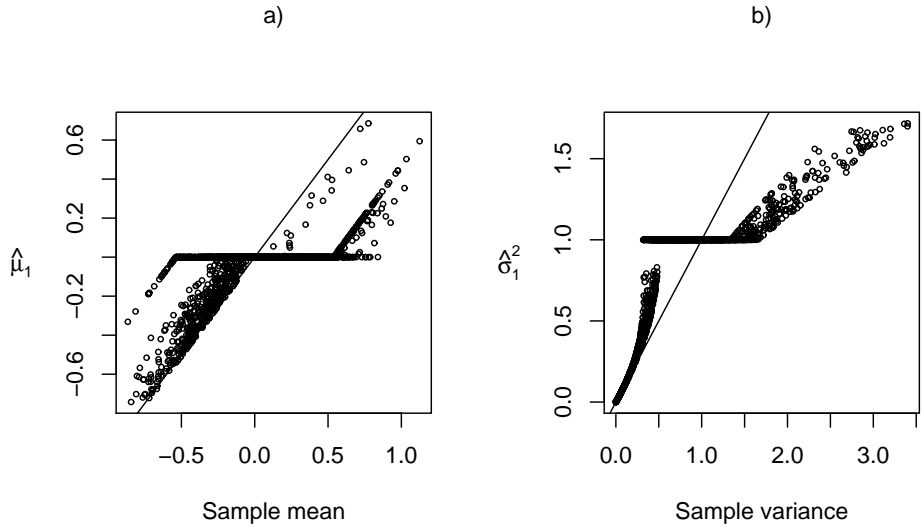

FIG 3. *Penalized mean and variance estimates for cluster one containing the 11 ALL B-cell samples by the new penalized method.*

Finally, we show the effects of shrinkage and thresholding for the parameter estimates by the new penalized method. Figure 3 depicts the penalized mean estimates (panel a) and variance estimates (panel b) versus the sample means and variances respectively for cluster one. It is confirmed that the penalized mean estimates were shrunk towards 0, some of which were exactly 0, while the penalized variance estimates were shrunk towards 1, and can be exactly 1.

Table 6
*Clustering results for Golub's data by PAM and K-means methods. The number of clusters selected by the silhouette width are masked by * *

| Methods | PAM | | | | | | | | | | | K-means | | | | | | | | | |
|---|---|---|---|---|---|---|---|---|---|---|---|---|---|---|---|---|---|---|---|---|---|
| RI/aRI | 0.46/0 | | 0.65/0.24 | | | 0.64/0.22 | | | | 0.67/0.27 | | | 0.80/0.53 | | | | 0.75/0.42 | | | | | |
| Clusters | 1 | 2* | 1 | 2 | 3 | 1 | 2 | 3 | 4 | 1 | 2 | 3 | 1 | 2 | 3 | 4 | 1 | 2 | 3 | 4 | 5 | 6* |
| Samples (#) | | | | | | | | | | | | | | | | | | | | | | |
| ALL-T (8) | 7 | 1 | 1 | 7 | 0 | 1 | 7 | 0 | 0 | 5 | 3 | 0 | 0 | 0 | 8 | 0 | 0 | 0 | 0 | 0 | 8 | 0 |
| ALL-B (19) | 11 | 8 | 8 | 3 | 8 | 8 | 3 | 7 | 1 | 10 | 7 | 2 | 10 | 2 | 0 | 7 | 1 | 9 | 0 | 8 | 0 | 1 |
| AML (11) | 11 | 0 | 11 | 0 | 0 | 11 | 0 | 9 | 0 | 0 | 1 | 10 | 0 | 10 | 0 | 1 | 0 | 4 | 7 | 0 | 0 | 0 |

### 4.2.2. Other clustering methods

Previous studies (e.g. Thalamuthu et al. 2006 [44]) have established model-based clustering as one of the best performers for gene expression data. Although it is not our main goal here, as a comparison in passing, we show the results of other three widely used methods as applied to the same data with the top 2000 genes: hierarchical clustering, partitioning around medoids (PAM) and K-means clustering.

It is challenging to determine the number of clusters for PAM and K-means. Here we consider two proposals. First, we used the silhouette width (Kaufman and Rousseeuw 1990 [24]) to select the number of clusters. Two and six clusters were chosen for PAM and K-means respectively; neither gave a good separation among the three leukemia subtypes (Table 6). Second, to sidestep the issue, we applied the two methods with three and four clusters because those numbers seemed to work best for model-based clustering. Nevertheless, PAM worked poorly, while K-means with 4 clusters gave a reasonable result, albeit not as good as that of the new penalized model-based clustering, as judged by an eye-ball examination or the (adjusted) Rand index.

Figure 4 gives the results of hierarchical clustering with all three ways of defining a cluster-to-cluster distance: average linkage, single linkage and complete linkage. The average linkage clustering gave the best separation among the three leukemia subtypes: all 8 T-cell samples, except sample 7, were grouped together; there were 10 B-cell samples in a group; all other ALL samples seemed to appear in the AML group. On the other hand, the average linkage clustering identified about six outlying samples, which were samples 7, 18, 19, 21, 22 and 27 respectively; this finding was consistent with that of the penalized model-based clustering with an equal covariance matrix, which detected the same five outliers except sample 19.

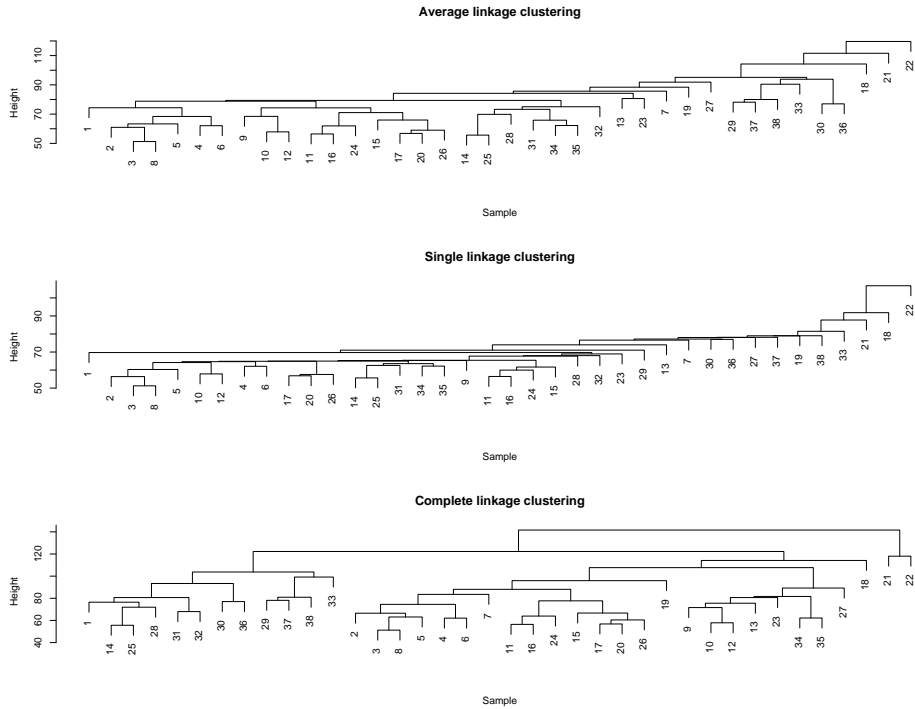

Fig 4. *Agglomerative hierarchical clustering results for the 38 leukemia samples: the first 8 samples were T-cell ALL; samples 9-27 were B-cell ALL; the remaining ones were AML.*

### 4.2.3. Other comparisons

Although mainly studied in the context of supervised learning, with several existing studies, Golub's data may serve as a test bed to compare various clustering methods. Golub et al. (1999) [15] applied self-organizing maps (SOM): first, with two clusters, SOM mis-classified one AML and 3 ALL samples; second, with four clusters, similar to the result of our new penalized method, AML and ALL-T each formed a cluster while ALL-B formed two clusters, in which one ALL-B and one AML samples were mis-classified. They did not discuss why or how $g=2$ or $g=4$ clusters were chosen. In Bayesian model-based clustering by Liu et al. (2003) [30], two clusters were chosen with one AML and one ALL mis-assigned; they did not discuss classification with ALL subtypes.

In a very recent study by Kim et al. (2006) [25], with two clustering algorithms and two choices of a prior parameter, they presented four sets of clustering results. In general, ALL samples formed one cluster while AML samples formed 5 to 6 clusters, giving 0-3 mis-assigned ALL samples; although not discuss explicitly, because either all or almost all the ALL samples fell into one cluster, their method obviously could not distinguish the two subtypes of ALL. Furthermore, their result on the multiple clusters of AML was in contrast to ours and Golub's on the homogeneity of AML samples. Because Kim et al. used different data pre-processing with 3571 genes as input to their method, for a fair comparison, we applied the same dataset to our new penalized method, yielding five clusters: only one ALL-B was mis-assigned to a cluster containing 10 AML samples, one cluster was consisted of one ALL-B and AML samples, while the other three clusters contained 8 ALL-T, 10 ALL-B and 7 ALL-B respectively. For this dataset, our method seemed to work better.

It was somewhat surprising that there were about 1800 genes remaining for the penalized methods, though previous studies showed that there were a large number of the genes differentially expressed between ALL and AML; in particular, Golub et al. (1999) [15] stated that "roughly 1100 genes were more highly correlated with the AML-ALL class distinction than would be expected by chance"; see also Pan and Shen (2007) [39] and references therein. For a simple evaluation, we applied the elastic net (Zou and Hastie 2005 [58]) to the same data with top 2000 genes; the elastic net is a state-of-the-art supervised learning method specifically designed for variable selection for high-dimensional data and is implemented in R package `elasticnet`. Five-fold cross-validation was used for tuning parameter selection. As usual, we decomposed the three-class problem into two binary classifiers, ALL-T vs AML, and ALL-B vs ALL, respectively. The two classifiers eliminated 395 and 870 noise genes, respectively, with a common set of 227 genes. Hence the elastic net used 1773 genes, a number comparable to those selected by the penalized clustering methods.

## *4.3. Grouping genes*

The 2000 genes were grouped according to the Kyoto Encyclopedia of Genes and Genomes (KEGG) Pathway database (Kanehisa and Goto 2000 [23]). About 43 percent of the 2000 genes belonged to at least one of 126 KEGG pathways. If a gene was in more than one pathway, it was randomly assigned to one of the pathways to which it belonged. Each genes un-annotated in any pathway was treated as an individual group with size 1. Among the 126 KEGG pathways, the largest pathway size was 44, the smallest one was 1 and the median size was 4; about three quarters of the pathways had sizes less than 8.

The clustering results with the grouped mean and variance penalization were exactly the same as that of UnequalCov and kept 1795 genes. Among the 205 identified noise genes, 23 genes were from 17 KEGG pathways: all contained only one genes except only three pathways, each containing 2, 3 and 4 genes respectively.

To further evaluate the above gene selection results, we searched a Leukemia Gene Database containing about 70 genes that were previously identified in the literature as leukemia-related (www.bioinformatics.org/legend/leuk_db.htm). Among these informative genes, 58 were related to 21 leukemia subtypes, among which only 47 and 36 genes appeared in the whole Golub's data and the 3337 genes after preprocessing respectively. Among the top 2000 genes being used for clustering, there were only 30 genes in the Leukemia Gene Database, most of which were not in any KEGG pathways; only 7 genes appeared in KEGG pathways: GMPS, ETS1, NOTCH3, MLL3, MYC, NFKB2 and KIT. Table 7 lists the genes that were selected in and deleted from the final model. Among the 205 noise genes selected by our group penalized method, five of them were annotated in the Leukemia Gene Database, among which one was related to AML.

TABLE 7
*The genes in the Leukemia Gene Database that were retained in or removed from the final model for the grouped method. The six genes in* italic *font were annotated in KEGG pathways*

|  | Leukemia Subtype | Gene Name |
| --- | --- | --- |
| Retained | Acute Lymphoblastic Leukemia | *MYC*, ZNFN1A1 |
|  | Acute Myelogenous Leukemia | IRF1, *GMPS* |
|  | Acute Myeloid Leukemia | CBFB, NUP214, HOXA9, FUS, RUNX1 |
|  | Acute Promyelocytic Leukemia | PML |
|  | Acute Undifferentiated Leukemia | SET |
|  | B-cell Chronic Lymphocytic Leukemia | BCL3, BTG1 |
|  | Myeloid Leukemia | CLC |
|  | pre B-cell Leukemia | PBX1, PBX3 |
|  | T-cell Leukemia | TCL6 |
|  | T-cell Acute Lymphoblastic Leukemia | *NOTCH3*, LYL1, LMO2, TAL2 |
|  | Cutaneous T-cell Leukemia | *NFKB2* |
|  | Human Monocytic Leukemia | *ETS1* |
|  | Mast cell Leukemia | *KIT* |
|  | Mixed Linkage Leukemia | *MLL3* |
| Removed | Acute Myeloid Leukemia | LCP1 |
|  | Acute Myelogenous Leukemia | RGS2 |
|  | Murine Myeloid Leukemia | EVI2B |
|  | pre B-cell Leukemia | PBX2 |
|  | T-cell Leukemia | TRA@ |

Because most of the known leukemia genes were not in any KEGG pathways, reflecting perhaps the current lack of prior knowledge, the grouped method could not be established as a clear winner over the none-grouped method in terms of leukemia gene selection in the above example. To confirm the potential gain with a better use of prior knowledge, we did two additional experiments. First, in addition to the KEGG pathways, we grouped all the 19 leukemia genes not in any KEGG pathways into a separate group: the samples were clustered as before; among the 200 genes removed from the final model, there were only two leukemia gene, ETS1, which was related to human monocytic leukemia, neither AML nor ALL, and NOTCH3, related to T-cell ALL. Second, in addition to the KEGG pathways, we grouped the AML ("acute myeloid leukemia" in Table 7) or ALL ("acute lymphoblastic leukemia" and "T-cell acute lymphoblastic leukemia") genes into two corresponding groups while treating the other leukemia genes individually: again the samples were clustered as before; among the 216 genes removed from the final model, ETS1, RGS2, EVI2B, PBX2, TRA@ were the four leukemia genes and there was no single gene related to AML or ALL. These two experiments demonstrated the effectiveness of grouping genes based on bi-ological knowledge, and that, as expected, the quality of the prior knowledge would influence performance. Nevertheless, our work here is just a first step, and more research is necessary to establish the practical use of grouping genes for microarray data.

## 5. Discussion

We have proposed a new penalized likelihood method for variable selection in model-based clustering, permitting cluster-dependent diagonal covariance matrices. A major novelty is the development of a new $L_1$ penalty involving both mean and variance parameters. The penalized mixture model can be fitted easily using an EM algorithm. Our numerical studies demonstrate the utility of the proposed method and its superior performance over other methods. In particular, it is confirmed that for high-dimensional data such as arising from microarray experiments, variable selection is necessary: without variable selection, the presence of a large number of noise variables can mask the clustering structure underlying the data. Furthermore, we have also studied penalties for grouped variables to incorporate prior knowledge into clustering analysis, which, as expected, improves performance if the prior knowledge being used is indeed informative.

The present approach involves only diagonal covariance matrices. It is argued that for "high dimension but small sample size" settings as arising in genomic studies, the working independence assumption is effective, as suggested by Fraley and Raftery (2006) [12], as well as demonstrated by the popular use of a diagonal covariance matrix in the naive Bayes and other discriminant analyses due to its good performance (Bickel and Levina 2004 [5]; Dudoit et al. 2002 [7]; Tibshirani et al. 2003 [47]). Nevertheless, it is worthwhile to generalize the proposed approach to other non-diagonal covariance matrices, possibly built on the novel idea of shrinking variance components as proposed here. However, this task is much more challenging; a main difficulty is how to guarantee a shrunk covariance matrix to be positive definite, as evidenced by the challenge in a simpler context of penalized estimation of a single covariance matrix (Huang et al. 2006 [21]; Yuan and Lin 2007 [55]). An alternative approach is to have a model intermediate between the independent and unrestricted models. For example, in a mixture of factor analyzers (McLachlan et al. 2003 [35]), local dimension reduction within each component is realized through some latent factors, which are also used to explain the correlations among the variables. Nevertheless, because the latent factors are assumed to be shared by all the variables while in fact they may only be related to a small subset of informative variables, variable selection may still be necessary; however, how to do so is an open question. Finally, although our proposed penalty for grouped variables provides a general framework to consider a group of genes, e.g. in a relevant biological pathway or functional group, for their either "all in" or "all out" property in clustering, there remain some practical questions, such as how to choose pathways and how to handle genes in multiple pathways. These interesting topics remain to be studied.

### Appendix A: Composite Absolute Penalties (CAP)

We generalize our proposed group penalization, including the two regularization schemes on variance parameters, to the Composite Absolute Penalties (CAP) of Zhao et al. (2006) [56], which covers the group penalty of Yuan and Lin (2006) [54] as a special case.

For grouping mean parameters, the following penalty function is used for the mean parameters:

$$\lambda_1 \left( \sum_{i=1}^{g} \sum_{m=1}^{M} k_m^{\gamma_0/\gamma_m'} \|\mu_i^m\|_{\gamma_m}^{\gamma_0} \right)^{1/\gamma_0} \quad (26)$$

where $1/\gamma_m + 1/\gamma_m' = 1$, $\gamma_m > 1$ and $\|\mathbf{v}\|_{\gamma_m}$ is the $L_{\gamma_m}$ norm of vector $\mathbf{v}$. Accordingly, we adopt

$$\lambda_2 \left( \sum_{i=1}^{g} \sum_{m=1}^{M} k_m^{\gamma_0/\gamma_m'} \|\sigma_{i,m}^2 - \mathbf{1}\|_{\gamma_m}^{\gamma_0} \right)^{1/\gamma_0} \quad (27)$$

or

$$\lambda_2 \left( \sum_{i=1}^{g} \sum_{m=1}^{M} k_m^{\gamma_0/\gamma_m'} \|\log \sigma_{i,m}^2\|_{\gamma_m}^{\gamma_0} \right)^{1/\gamma_0} \quad (28)$$

as a penalty for grouped variance parameters. To achieve sparcity, as usual, we use $\gamma_0 = 1$.

The E-step of the EM yields $Q_P$ with the same form as (2). Next we derive the updating formulas for the mean and variance parameters in the M-step.

#### A.1. Grouping mean parameters

If the CAP penalty function (26) for grouped means is used, we can derive the following Theorem:

**Theorem 6.** *The sufficient and necessary conditions for $\mu = (\mu_i^m)$, $i = 1, 2, \ldots, g$ and $m = 1, 2, \ldots, M$, to be a unique maximizer of $Q_P$ are*

$$V_{im}^{-1} \left( \sum_{j=1}^{n} \tau_{ij} x_j^m - \left( \sum_{j=1}^{n} \tau_{ij} \right) \mu_i^m \right) = \lambda_1 k_m^{1/\gamma_m'} \frac{sign(\mu_i^m)|\mu_i^m|^{\gamma_m - 1}}{\|\mu_i^m\|_{\gamma_m}^{\gamma_m - 1}},$$
$$\text{if and only if } \mu_i^m \neq \mathbf{0}, \quad (29)$$

*and*

$$\left\| \sum_{j=1}^{n} \tau_{ij} x_j^{m\prime} V_{im}^{-1} \right\|_{\gamma_m'} \leq \lambda_1 k_m^{1/\gamma_m'}, \quad \text{if and only if } \mu_i^m = \mathbf{0}, \quad (30)$$

*yielding*

$$\hat{\mu}_i^m = \left( sign \left( 1 - \frac{\lambda_1 k_m^{1/\gamma_m'}}{\|\sum_{j=1}^{n} \tau_{ij} x_j^{m\prime} V_{im}^{-1}\|_{\gamma_m'}} \right) \right)_+ \nu_i^m \tilde{\mu}_i^m \quad (31)$$

where $\nu_i^m = \left(\mathbf{I} + \frac{\lambda_1 k_m^{1/\gamma_m'} \|\hat{\mu}_i^m\|^{\gamma_m - 2}}{\sum_{j=1}^n \tau_{ij} \|\hat{\mu}_i^m\|_{\gamma_m}} V_{im}\right)^{-1}$, and $\tilde{\mu}_i^m = \sum_{j=1}^n \tau_{ij} x_j^m / \sum_{j=1}^n \tau_{ij}$ is the usual MLE.

*Proof.* Consider two cases:

i) $\mu_i^m \neq \mathbf{0}$. First, by definition and using the Hölder's inequality, we can prove that the $L_{\gamma_m}$ norm is convex, thus the penalty function for grouped means is convex in $\mu_i^m$. Second, treating $Q_P$ as the Lagrange multiplier for a constrained optimization problem with the penalty as the inequality constraint, and considering that both minus the objective function and the penalty function are convex, by the Karush-Kuhn-Tucker (KKT) condition, we have the following sufficient and necessary condition

$$\partial Q_P / \partial \mu_i^m = \mathbf{0} \iff \sum_j \tau_{ij} V_{im}^{-1}(x_j^m - \mu_i^m) - \lambda_1 k_m^{1/\gamma_m'} \frac{\text{sign}(\mu_i^m)|\mu_i^m|^{\gamma_m - 1}}{\|\mu_i^m\|_{\gamma_m}^{\gamma_m - 1}} = \mathbf{0},$$

from which we can easily get (29).

ii) $\mu_i^m = \mathbf{0}$. By definition, we have

$$Q_P(\mathbf{0}, .) \geq Q_P(\Delta \mu_i^m, .) \text{ for any } \Delta \mu_i^m \text{ close to } \mathbf{0}$$

$$\iff -\sum_j \tau_{ij} \frac{1}{2} (x_j^m)' V_{im}^{-1} x_j^m + C_1 \geq$$

$$-\sum_j \tau_{ij} \frac{1}{2} (x_j^m - \Delta \mu_i^m)' V_{im}^{-1} (x_j^m - \Delta \mu_i^m) - \lambda_1 k_m^{1/\gamma_m'} \|\Delta \mu_i^m\|_{\gamma_m} + C_1$$

$$\iff \sum_j \tau_{ij} x_j^{m'} V_{im}^{-1} \Delta \mu_i^m / \|\Delta \mu_i^m\|_{\gamma_m} - \sum_j \tau_{ij} (\Delta \mu_i^m)' V_{im}^{-1} \Delta \mu_i^m / (2\|\Delta \mu_i^m\|_{\gamma_m})$$

$$\leq \lambda_1 k_m^{1/\gamma_m'}.$$

Notice $\sum_j \tau_{ij} (\Delta \mu_i^m)' V_{im}^{-1} \Delta \mu_i^m / (2\|\Delta \mu_i^m\|_{\gamma_m}) \to 0^+$ as $\Delta \mu_i^m \to \mathbf{0}$. By the Hölder's inequality, we have $\left|\sum_j \tau_{ij} x_j^{m'} V_{im}^{-1} \Delta \mu_i^m / \|\Delta \mu_i^m\|_{\gamma_m}\right| \leq \|\sum_j \tau_{ij} x_j^{m'} V_{im}^{-1}\|_{\gamma_m'}$, and the " = " can be attained. Thus the above inequality is equivalent to (30). □

It is clear that, if $\lambda_1$ is large enough, $\hat{\mu}_i^m$ will be exactly $\mathbf{0}$ due to thresholding. Since $\nu_i^m$ depends on $\hat{\mu}_i^m$, we use (31) iteratively to update $\hat{\mu}_i^m$.

## *A.2. Grouping variance parameters*

### *A.2.1. Scheme 1*

If the penalty function (27) for grouped variances is used, we have the following theorem:

**Theorem 7.** *The sufficient and necessary conditions for $\sigma_{i,m}^2 = \mathbf{1}$, $i = 1, 2, \ldots, g$ and $m = 1, 2, \ldots, M$, to be a local maximizer of $Q_P$ are*

$$\begin{cases} \left\| \sum_{j=1}^n \tau_{ij} \left[ \frac{1}{2}\mathbf{1} - \frac{1}{2}(x_j^m - \mu_i^m)^2 \right] \right\|_{\gamma_m'} \leq \lambda_2 k_m^{1/\gamma_m'} & \text{if } \sum_{j=1}^n \tau_{ij} \left[ \frac{1}{2}\mathbf{1} - (x_j^m - \mu_i^m)^2 \right] \leq \mathbf{0}; \\ \left\| \sum_{j=1}^n \tau_{ij} \left[ \frac{1}{2}\mathbf{1} - \frac{1}{2}(x_j^m - \mu_i^m)^2 \right] \right\|_{\gamma_m'} < \lambda_2 k_m^{1/\gamma_m'} & \text{otherwise}. \end{cases}$$

(32)

*The necessary condition for $\sigma_{i,m}^2 \neq \mathbf{1}$ to be a local maximizer of $Q_P$ is*

$$\sum_{j=1}^n \tau_{ij} \left[ -\frac{1}{2\sigma_{i,m}^2} + \frac{1}{2(\sigma_{i,m}^2)^2}(x_j^m - \mu_i^m)^2 \right] = \frac{\lambda_2 k_m^{1/\gamma_m'} sign(\sigma_{i,m}^2 - \mathbf{1})|\sigma_{i,m}^2 - \mathbf{1}|^{\gamma_m - 1}}{\|\sigma_{i,m}^2 - \mathbf{1}\|_{\gamma_m}^{\gamma_m - 1}}.$$

(33)

*Proof.* If $\sigma_{i,m}^2 = \mathbf{1}$ is a local maximum, by definition, we have the following sufficient and necessary condition

$$Q_P(\mathbf{1},.) \geq Q_P(\mathbf{1} + \Delta\sigma_{i,m}^2,.) \text{ for any } \Delta\sigma_{i,m}^2 \text{ near } \mathbf{0}$$

$$\iff \sum_j \tau_{ij} \left[ -\frac{1}{2}(x_j^m - \mu_i^m)'(x_j^m - \mu_i^m) \right] + C_1 \geq$$

$$\sum_j \tau_{ij} \left[ -\frac{1}{2} \log|\text{diag}(\mathbf{1} + \Delta\sigma_{i,m}^2)| \right.$$

$$\left. - \frac{1}{2}(x_j^m - \mu_i^m)'\text{diag}(\mathbf{1} + \Delta\sigma_{i,m}^2)^{-1}(x_j^m - \mu_i^m) \right]$$

$$- \lambda_2 k_m^{1/\gamma_m'} \|\Delta\sigma_{i,m}^2\|_{\gamma_m} + C_1.$$

Thus,

$$\sum_j \tau_{ij} \left[ -\frac{1}{2} \log|\text{diag}(\mathbf{1} + \Delta\sigma_{i,m}^2)| \right.$$

$$\left. + \frac{1}{2}(x_j^m - \mu_i^m)'\text{diag}(\Delta\sigma_{i,m}^2/(\mathbf{1} + \Delta\sigma_{i,m}^2))(x_j^m - \mu_i^m) \right]$$

$$\leq \lambda_2 k_m^{1/\gamma_m'} \|\Delta\sigma_{i,m}^2\|_{\gamma_m}.$$

Using Taylor's expansion, we have

$$\sum_{j=1}^n \tau_{ij} \left( -\frac{1}{2}\mathbf{1} + \frac{(x_j^m - \mu_i^m)^2}{2} \right)' \Delta\sigma_{i,m}^2 + \frac{1}{2} \sum_{j=1}^n \tau_{ij} \left( \frac{1}{2}\mathbf{1} - (x_j^m - \mu_i^m)^2 \right)' (\Delta\sigma_{i,m}^2)^2$$

$$+ O\left(\mathbf{c}'(\Delta\sigma_{i,m}^2)^3\right) \leq \lambda_2 k_m^{1/\gamma_m'} \|\Delta\sigma_{i,m}^2\|_{\gamma_m}$$

for some constant vector $\mathbf{c}$. After dividing both sides by $\|\Delta\sigma_{i,m}^2\|_{\gamma_m}$ and using the same argument as before, we obtain (32) as the sufficient and necessary condition for $\sigma_{i,m}^2 = \mathbf{1}$ to be a local maximizer of $Q_P$.

Setting the first-order derivative of $Q_P$ equal to 0, we have (33), the necessary condition for $\sigma_{i,m}^2 \neq \mathbf{1}$ to be a local maximizer of $Q_P$. □

It is clear that we have $\sigma_{i,m}^2 = \mathbf{1}$ when, for example, $\lambda_2$ is large enough. It is also easy to verify that the above conditions reduce to the same ones for $\sigma_{ik}^2 = 1$ for non-grouped variables when $k_m = 1$ and reduce to (24) and (25) for grouped variables when $\gamma_m = \gamma_m' = 2$.

*A.2.2. Scheme 2*

If we use the CAP penalty function (28) for grouped variances, then the following theorem can be obtained by a similar argument as before:

**Theorem 8.** *The sufficient and necessary conditions for $\sigma_{i,m}^2 = \mathbf{1}$, $i = 1, 2, \ldots, g$ and $m = 1, 2, \ldots, M$, to be a local maximizer of $Q_P$ are*

$$\begin{cases} \left\| \sum_{j=1}^{n} \tau_{ij} \left[ \frac{1}{2}\mathbf{1} - \frac{1}{2}(x_j^m - \mu_i^m)^2 \right] \right\|_{\gamma_m'} \leq \lambda_2 k_m^{1/\gamma_m'} & \text{if } \sum_{j=1}^{n} \tau_{ij} \left[ \frac{1}{2}\mathbf{1} - (x_j^m - \mu_i^m)^2 \right] \leq \mathbf{0}; \\ \left\| \sum_{j=1}^{n} \tau_{ij} \left[ \frac{1}{2}\mathbf{1} - \frac{1}{2}(x_j^m - \mu_i^m)^2 \right] \right\|_{\gamma_m'} < \lambda_2 k_m^{1/\gamma_m'} & \text{otherwise.} \end{cases}$$

(34)

*The necessary condition for $\sigma_{i,m}^2 \neq \mathbf{1}$ to be a local maximizer of $Q_P$ is*

$$\sum_{j=1}^{n} \tau_{ij} \left[ -\frac{1}{2\sigma_{i,m}^2} + \frac{1}{2(\sigma_{i,m}^2)^2}(x_j^m - \mu_i^m)^2 \right] = \frac{\lambda_2 k_m^{1/\gamma_m'} sign(\log \sigma_{i,m}^2) |\log \sigma_{i,m}^2|^{\gamma_m - 1}}{\| \log \sigma_{i,m}^2 \|_{\gamma_m}^{\gamma_m - 1}}.$$

(35)

**Appendix B: Proofs**

*B.1. Derivation of Theorem 1*

Since $Q_P$ is differentiable with respect to $\mu_{ik}$ when $\mu_{ik} \neq 0$, while non-differentiable at $\mu_{ik} = 0$, we consider the following two cases:

i) If $\mu_{ik} \neq 0$ is a maximum, given that $Q_P$ is concave and differentiable, then the sufficient and necessary condition for $\mu_{ik}$ to be the global maximum of $Q_P$ is

$$\partial Q_P / \partial \mu_{ik} = 0 \iff \sum_{j=1}^{n} \tau_{ij}(x_{jk} - \mu_{ik})/\sigma_{ik}^2 - \lambda_1 \text{sign}(\mu_{ik}) = 0,$$

from which we have (10).

ii) If $\mu_{ik} = 0$ is a maximum, we compare $Q_P(0, .)$ with $Q_P(\Delta \mu_{ik}, .)$, the values of $Q_P$ at $\mu_{ik} = 0$ and $\mu_{ik} = \Delta \mu_{ik}$ respectively (while other components of $\mu_i$ are fixed at its maximum). By definition, we have

$$Q_P(0,.) \geq Q_P(\Delta\mu_{ik},.) \text{ for any } \Delta\mu_{ik} \text{ near } 0$$
$$\iff -\sum_j \tau_{ij}\frac{1}{2}x_{jk}^2/\sigma_{ik}^2 + C_1 \geq -\sum_j \tau_{ij}\frac{1}{2}(x_{jk}-\Delta\mu_{ik})^2/\sigma_{ik}^2 - \lambda_1|\Delta\mu_{ik}| + C_1$$
$$\iff \sum_j \tau_{ij}\frac{1}{2}(2x_{jk}\text{sign}(\Delta\mu_{ik}) - |\Delta\mu_{ik}|)/\sigma_{ik}^2 \leq \lambda_1$$
$$\iff \left|\sum_j \tau_{ij}x_{jk}\right|/\sigma_{ik}^2 \leq \lambda_1 \text{ as } \Delta\mu_{ik} \to 0.$$

It is obvious that from (10) we have $\text{sign}(\mu_{ik}) = \text{sign}(\sum_{j=1}^n \tau_{ij}x_{jk}/\sum_{j=1}^n \tau_{ij})$, thus

$$\mu_{ik} = \frac{\sum_{j=1}^n \tau_{ij}x_{jk}}{\sum_{j=1}^n \tau_{ij}}\left(1 - \frac{\lambda_1\sigma_{ik}^2\text{sign}(\mu_{ik})}{\sum_{j=1}^n \tau_{ij}x_{jk}}\right) = \frac{\sum_{j=1}^n \tau_{ij}x_{jk}}{\sum_{j=1}^n \tau_{ij}}\left(1 - \frac{\lambda_1\sigma_{ik}^2}{|\sum_{j=1}^n \tau_{ij}x_{jk}|}\right),$$

which, in combination with (11), yields (12).

### B.2. Derivation of Theorem 2

Since $Q_P$ is differentiable with respect to $\sigma_{ik}^2$ when $\sigma_{ik}^2 \neq 1$, we know a local maximum $\hat{\sigma}_{ik}^2$ must satisfy the following conditions

$$\begin{cases} \frac{\partial}{\partial\sigma_{ik}^2}Q_P(\Theta;\Theta^{(r)})|_{\sigma_{ik}^2=\hat{\sigma}_{ik}^2} = 0 & \text{if } \hat{\sigma}_{ik}^2 \neq 1; \\ Q_P(1,.) \geq Q_P(1+\Delta\sigma_{ik}^2,.) & \text{if } \hat{\sigma}_{ik}^2 = 1 \text{ and for any } \Delta\sigma_{ik}^2 \text{ near } 0. \end{cases} \quad (36)$$

where . in $Q_P(1,.)$ represents all parameters in $Q_P$ except $\sigma_{ik}^2$.

Notice that $Q_P = C_1 + \sum_j \tau_{ij}\left[-\frac{1}{2}\log\sigma_{ik}^2 + C_2 - \frac{1}{2}(x_{jk}-\mu_{ik})^2/\sigma_{ik}^2\right] - \lambda_2|\sigma_{ik}^2 - 1| + C_3$, where $C_1$, $C_2$ and $C_3$ are constants with respect to $\sigma_{ik}^2$. Therefore the first equation of (36) becomes

$$\sum_{j=1}^n \tau_{ij}\left(-\frac{1}{2\hat{\sigma}_{ik}^2} + \frac{(x_{jk}-\mu_{ik})^2}{2\hat{\sigma}_{ik}^4}\right) - \lambda_2\text{sign}(\hat{\sigma}_{ik}^2 - 1) = 0, \quad \text{if } \hat{\sigma}_{ik}^2 \neq 1$$

from which we can easily get (13).

Starting from the second equation of (36), we have

$$\text{LHS} = C_1 + \sum_j \tau_{ij}\left[-\frac{1}{2}\log(1) + C_2 - \frac{1}{2}(x_{jk}-\mu_{ik})^2/1\right] - \lambda_2|1-1| + C_3,$$

$$\text{RHS} = C_1 + \sum_j \tau_{ij}\left[-\frac{1}{2}\log(1+\Delta\sigma_{ik}^2) + C_2 - \frac{1}{2}(x_{jk}-\mu_{ik})^2/(1+\Delta\sigma_{ik}^2)\right]$$
$$- \lambda_2|\Delta\sigma_{ik}^2| + C_3,$$

and thus

$$\frac{1}{2}\sum_j \tau_{ij}\left[-\log(1+\Delta\sigma_{ik}^2)-(x_{jk}-\mu_{ik})^2(1/(1+\Delta\sigma_{ik}^2)-1)\right] \le \lambda_2|\Delta\sigma_{ik}^2|.$$

Using Taylor's expansion, we have

$$\sum_{j=1}^n \tau_{ij}\left(-\frac{1}{2}+\frac{(x_{jk}-\mu_{ik})^2}{2}\right)\Delta\sigma_{ik}^2 + O((\Delta\sigma_{ik}^2)^2) \le \lambda_2|\Delta\sigma_{ik}^2|,$$

leading to

$$\sum_{j=1}^n \tau_{ij}\left(-\frac{1}{2}+\frac{(x_{jk}-\mu_{ik})^2}{2}\right)\text{sign}(\Delta\sigma_{ik}^2) + O(|\Delta\sigma_{ik}^2|) \le \lambda_2.$$

letting $\Delta\sigma_{ik}^2 \to 0$, we obtain (14).

### B.3. Derivation of $\hat{\sigma}_{ik}^2$ in section 2.3.1

Note that from (13) we have $(-\partial Q_P/\partial\sigma_{ik}^2)\sigma_{ik}^4 = a_{ik}\sigma_{ik}^4 + b_i\sigma_{ik}^2 - c_{ik} = 0$. Define $f(x) = a_{ik}x^2 + b_i x - c_{ik} = 0$.

First, we consider the case with $|b_i - c_{ik}| \le \lambda_2$.

i) When $\tilde{\sigma}_{ik}^2 > 1$, $f(\tilde{\sigma}_{ik}^2) = a_{ik}\tilde{\sigma}_{ik}^4 + 0 = \lambda_2\tilde{\sigma}_{ik}^4 > 0$, and $f(x) = \lambda_2\text{sign}(x-1)x^2 + b_i x - c_{ik} = \lambda_2 x^2 + b_i x - c_{ik} > b_i x - c_{ik} \ge b_i\tilde{\sigma}_{ik}^2 - c_{ik} = 0$ if $x > \tilde{\sigma}_{ik}^2 > 1$. On the other hand, $\lim_{x\to 1^+} f(x) = \lambda_2 + b_i - c_{ik} \ge 0$, since $|b_i - c_{ik}| \le \lambda_2$; and $f(x) = -\lambda_2 x^2 + b_i x - c_{ik} < -\lambda_2 x^2 + b_i - c_{ik} < 0$ if $x < 1$. Thus, based on the signs of $f(x)$, $Q_P$ has a unique local maximum at $\hat{\sigma}_{ik}^2 = 1$.

ii) When $\tilde{\sigma}_{ik}^2 < 1$, we have $f(\tilde{\sigma}_{ik}^2) = -\lambda_2\tilde{\sigma}_{ik}^4 < 0$, and $f(x) = -\lambda_2 x^2 + b_i x - c_{ik} < b_i x - c_{ik} < b_i\tilde{\sigma}_{ik}^2 - c_{ik} = 0$ if $x < \tilde{\sigma}_{ik}^2$; $\lim_{x\to 1^-} f(x) = -\lambda_2 + b_i - c_{ik} \le 0$; and $f(x) = \lambda_2 x^2 + b_i x - c_i > \lambda_2 + b_i - c_i > 0$ if $x > 1$. However, for $\tilde{\sigma}_{ik}^2 < x < 1$, $f(x) = -\lambda_2 x^2 + b_i x - c_{ik}$ is a continuous and quadratic function, which may have two roots

$$x_{1,2} = \frac{b_i \pm \sqrt{b_i^2 - 4\lambda_2 c_{ik}}}{2\lambda_2}.$$

If $b_i - c_{ik} < \lambda_2$, then $\lim_{x\to 1^-} f(x) < 0$, implying that, according to the signs of $f(x)$ around $x = 1$, $x = 1$ is a local maximum of $Q_P$, and the smaller of $x_{1,2}$ is also a local maximum (if it exists); on the other hand, if $b_i - c_{ik} = \lambda_2$, then $\lim_{x\to 1^-} f(x) = 0$, implying that either $x = 1$, the smaller root of $x_{1,2}$ if $\tilde{\sigma}_{ik}^2 \ge 1/2$, or $x = c_{ik}/\lambda_2$, the larger root of $x_{1,2}$ if $\tilde{\sigma}_{ik}^2 < 1/2$, is the unique maximum.

Second, we claim that, if $|b_i - c_{ik}| > \lambda_2$, there exists a unique local maximizer $\hat{\sigma}_{ik}^2 \ne 1$ for $Q_P$ and it must lie between 1 and $\tilde{\sigma}_{ik}^2 = c_{ik}/b_i$, the usual MLE without penalty. This can be shown in the following way.

i) When $\tilde{\sigma}_{ik}^2 > 1$, $f(\tilde{\sigma}_{ik}^2) = a_{ik}\tilde{\sigma}_{ik}^4 + 0 = \lambda_2\tilde{\sigma}_{ik}^4 > 0$, and $f(x) = \lambda_2\text{sign}(x-1)x^2 + b_i x - c_{ik} = \lambda_2 x^2 + b_i x - c_{ik} > b_i x - c_{ik} \ge b_i\tilde{\sigma}_{ik}^2 - c_{ik} = 0$ if $x > \tilde{\sigma}_{ik}^2 > 1$.

On the other hand, $\lim_{x \to 1^+} f(x) = \lambda_2 + b_i - c_{ik} < 0$, since $b_i - c_{ik} < -\lambda_2$; and $f(x) = -\lambda_2 x^2 + b_i x - c_{ik} < -\lambda_2 x^2 + b_i - c_{ik} < 0$ if $x < 1$. Thus $f(x)$ has a unique root $\hat{\sigma}_{ik}^2 \in (1, \tilde{\sigma}_{ik}^2)$.

$ii$) When $\tilde{\sigma}_{ik}^2 < 1$, similarly we have $f(\tilde{\sigma}_{ik}^2) < 0$, and $f(x) = -\lambda_2 x^2 + b_i x - c_{ik} < b_i x - c_{ik} < b_i \tilde{\sigma}_{ik}^2 - c_{ik} = 0$ if $x < \tilde{\sigma}_{ik}^2$; $\lim_{x \to 1^-} f(x) = -\lambda_2 + b_i - c_{ik} > 0$; and $f(x) = \lambda_2 x^2 + b_i x - c_i > b_i - c_i > 0$ if $x > 1$. Thus $f(x)$ has a unique root $\hat{\sigma}_{ik}^2 \in (\tilde{\sigma}_{ik}^2, 1)$.

Based on the signs of $f(x)$ around $x = \hat{\sigma}_{ik}^2$, it is easy to see that $\hat{\sigma}_{ik}^2$ is indeed a local maximizer.

Third, (13) can be expressed as

$$\begin{cases} -\lambda_2 \sigma_{ik}^4 + b_i \sigma_{ik}^2 - c_{ik} = 0, & \text{if } b_i - c_{ik} > \lambda_2 \\ \lambda_2 \sigma_{ik}^4 + b_i \sigma_{ik}^2 - c_{ik} = 0, & \text{if } b_i - c_{ik} < -\lambda_2. \end{cases} \quad (37)$$

From the first equation of (37), we get $\hat{\sigma}_{ik}^2 = \left(b_i \pm \sqrt{b_i^2 - 4\lambda_2 c_{ik}}\right)/2\lambda_2$. Since $b_i + \sqrt{b_i^2 - 4\lambda_2 c_{ik}} > b_i + \sqrt{(c_{ik} + \lambda_2)^2 - 4\lambda_2 c_{ik}} > (c_{ik} + \lambda_2) + |c_{ik} - \lambda_2| > 2\lambda_2$ and that $b_i - c_{ik} > \lambda_2$ implies $\tilde{\sigma}_{ik}^2 = c_{ik}/b_i < 1$ while $\hat{\sigma}_{ik}^2$ must be between $\tilde{\sigma}_{ik}^2$ and 1, we only have one solution $\hat{\sigma}_{ik}^2 = \left(b_i - \sqrt{b_i^2 - 4\lambda_2 c_{ik}}\right)/2\lambda_2 = \tilde{\sigma}_{ik}^2 / \left(\frac{1}{2} + \sqrt{\frac{1}{4} - \frac{\lambda_2 c_{ik}}{b_i^2}}\right)$. From the second equation, similarly we get $\hat{\sigma}_{ik}^2 = \tilde{\sigma}_{ik}^2 / \left(\frac{1}{2} + \sqrt{\frac{1}{4} + \frac{\lambda_2 c_{ik}}{b_i^2}}\right)$. Combining the two cases, we obtain (15).

Note that the expression inside the square root of (15) is non-negative. To prove it, we only need to show that for $b_i^2 + 4\text{sign}(c_{ik} - b_i)\lambda_2 c_{ik}$. Consider two cases:

$i$) When $c_{ik} - b_i > \lambda_2 \geq 0$,

$$b_i^2 + 4\text{sign}(c_{ik} - b_i)\lambda_2 c_{ik} = b_i^2 + 4\lambda_2 c_{ik} > b_i^2 + 4\lambda_2(b_i + \lambda_2) = (b_i + 2\lambda_2)^2 \geq 0.$$

$ii$) When $c_{ik} - b_i < -\lambda_2 \leq 0$,

$$b_i^2 + 4\text{sign}(c_{ik} - b_i)\lambda_2 c_{ik} = b_i^2 - 4\lambda_2 c_{ik} > b_i^2 - 4\lambda_2(b_i - \lambda_2) = (b_i - 2\lambda_2)^2 \geq 0.$$

### *B.4. Derivation of Theorem 3*

We prove the necessary conditions below, while the sufficiency is proved as a side-product in Appendix B.5.

Since $Q_P$ is differentiable with respect to $\sigma_{ik}^2$ when $\sigma_{ik}^2 \neq 1$, we know a local maximum $\hat{\sigma}_{ik}^2$ must satisfy the following conditions

$$\begin{cases} \left.\dfrac{\partial}{\partial \sigma_{ik}^2} Q_P(\Theta; \Theta^{(r)})\right|_{\sigma_{ik}^2 = \hat{\sigma}_{ik}^2} = 0 & \text{if } \hat{\sigma}_{ik}^2 \neq 1; \\ Q_P(1, .) \geq Q_P(1 + \Delta \sigma_{ik}^2, .) & \text{if } \hat{\sigma}_{ik}^2 = 1 \text{ and for any } \Delta \sigma_{ik}^2 \text{ near } 0. \end{cases} \quad (38)$$

where . in $Q_P(1,.)$ represents all parameters in $Q_P$ except $\sigma_{ik}^2$.

Notice that $Q_P = C_1 + \sum_j \tau_{ij} \left[ -\frac{1}{2} \log \sigma_{ik}^2 + C_2 - \frac{1}{2}(x_{jk} - \mu_{ik})^2/\sigma_{ik}^2 \right] - \lambda_2 |\log \sigma_{ik}^2| + C_3$, where $C_1$, $C_2$ and $C_3$ are constants with respect to $\sigma_{ik}^2$. Therefore the first equation of (38) becomes

$$\sum_{j=1}^n \tau_{ij} \left( -\frac{1}{2\hat{\sigma}_{ik}^2} + \frac{(x_{jk} - \mu_{ik})^2}{2\hat{\sigma}_{ik}^4} \right) - \frac{\lambda_2 \text{sign}(\log \hat{\sigma}_{ik}^2)}{\hat{\sigma}_{ik}^2} = 0, \quad \text{if } \hat{\sigma}_{ik}^2 \neq 1$$

from which we can easily get (17).

Starting from the second equation of (38), we have

$$\text{LHS} = C_1 + \sum_j \tau_{ij} \left[ -\frac{1}{2}\log(1) + C_2 - \frac{1}{2}(x_{jk} - \mu_{ik})^2/1 \right] - \lambda_2 |\log 1| + C_3,$$

$$\text{RHS} = C_1 + \sum_j \tau_{ij} \left[ -\frac{1}{2}\log(1 + \Delta\sigma_{ik}^2) + C_2 - \frac{1}{2}(x_{jk} - \mu_{ik})^2/(1 + \Delta\sigma_{ik}^2) \right]$$
$$- \lambda_2 |\log(1 + \Delta\sigma_{ik}^2)| + C_3,$$

and thus

$$\frac{1}{2}\sum_j \tau_{ij} \left[ -\log(1 + \Delta\sigma_{ik}^2) - (x_{jk} - \mu_{ik})^2 (1/(1 + \Delta\sigma_{ik}^2) - 1) \right] \leq \lambda_2 |\log(1 + \Delta\sigma_{ik}^2)|.$$

Using Taylor's expansion, we have

$$\sum_{j=1}^n \tau_{ij} \left( -\frac{1}{2} + \frac{(x_{jk} - \mu_{ik})^2}{2} \right) \Delta\sigma_{ik}^2 + O((\Delta\sigma_{ik}^2)^2) \leq \lambda_2 |\log(1 + \Delta\sigma_{ik}^2)|,$$

leading to

$$\sum_{j=1}^n \tau_{ij} \left( -\frac{1}{2} + \frac{(x_{jk} - \mu_{ik})^2}{2} \right) \text{sign}(\Delta\sigma_{ik}^2) + O(|\Delta\sigma_{ik}^2|) \leq \lambda_2 \left| \frac{\log(1 + \Delta\sigma_{ik}^2)}{\Delta\sigma_{ik}^2} \right|.$$

letting $\Delta\sigma_{ik}^2 \to 0$, we obtain (18).

## B.5. Derivation of $\hat{\sigma}_{ik}^2$ in section 2.3.2

Let $f(\sigma_{ik}^2) = -2\sigma_{ik}^4(\partial Q_P/\partial\sigma_{ik}^2) = \sigma_{ik}^2[1 + \lambda_2\text{sign}(\log\sigma_{ik}^2)/b_i] - \tilde{\sigma}_{ik}^2$, where $f(x)$ is defined as $f(x) = x[1 + \lambda_2\text{sign}(\log x)/b_i] - \tilde{\sigma}_{ik}^2$. Thus (17) is equivalent to $f(\hat{\sigma}_{ik}^2) = 0$. First, we consider the case with $|b_i - c_{ik}| \leq \lambda_2$, the necessary condition of $\hat{\sigma}_{ik}^2 = 1$.

i) When $\tilde{\sigma}_{ik}^2 > 1$, $f(x) = x[1 + \lambda_2/b_i] - \tilde{\sigma}_{ik}^2 > 1 + \lambda_2/b_i - c_{ik}/b_i = [(b_i - c_{ik}) + \lambda_2]/b_i > 0$ if $x > 1$. On the other hand, $f(x) = x[1 - \lambda_2/b_i] - \tilde{\sigma}_{ik}^2$ $(x - \tilde{\sigma}_{ik}^2) - x\lambda_2/b_i < -x\lambda_2/b_i < 0$ if $x < 1 < \tilde{\sigma}_{ik}^2$. Thus, based on the signs of $f(x)$, $Q_P$ has a unique local maximum at $\hat{\sigma}_{ik}^2 = 1$.

ii) When $\tilde{\sigma}_{ik}^2 < 1$, we have $c_{ik}/b_i < 1$, thus $0 < b_i - c_{ik} < \lambda_2$. $f(x) = x[1 + \lambda_2/b_i] - \tilde{\sigma}_{ik}^2 = (x - \tilde{\sigma}_{ik}^2) + x\lambda_2/b_i > 0$ if $x > 1$. On the other hand, $f(x) = x[1 - \lambda_2/b_i] - \tilde{\sigma}_{ik}^2 = x[b_i - \lambda_2]/b_i - \tilde{\sigma}_{ik}^2 < xc_{ik}/b_i - \tilde{\sigma}_{ik}^2 = (x-1)\tilde{\sigma}_{ik}^2 < 0$ if $0 < x < 1$. Thus, based on the signs of $f(x)$, $Q_P$ has a unique local maximum at $\hat{\sigma}_{ik}^2 = 1$.

$i)$ and $ii)$ indicates that $|b_i - c_{ik}| \le \lambda_2$ is also the sufficient condition of $\hat{\sigma}_{ik}^2 = 1$

Second, we claim that, if $|b_i - c_{ik}| > \lambda_2$, there exists a unique local maximizer $\hat{\sigma}_{ik}^2 \ne 1$ for $Q_P$ and it must lie between 1 and $\tilde{\sigma}_{ik}^2 = c_{ik}/b_i$, the usual MLE without penalty. This can be shown in the following way.

$i)$ When $\tilde{\sigma}_{ik}^2 > 1$, we have $b_i < c_{ik}$, and further $c_{ik} - b_i > \lambda_2$. Notice $f(x) = x[1 - \lambda_2/b_i] - \tilde{\sigma}_{ik}^2 = (x - \tilde{\sigma}_{ik}^2) - x\lambda_2/b_i < 0$ if $0 < x < 1$. Thus the possible root of $f(x) = 0$ should be larger than 1. For $x > 1$, $f(x) = x[1 + \lambda_2/b_i] - \tilde{\sigma}_{ik}^2$ is a linear function of $x$. $f(\tilde{\sigma}_{ik}^2) = \lambda_2 \tilde{\sigma}_{ik}^2/b_i > 0$, and $\lim_{x \to 1^+} f(x) = [1 + \lambda_2/b_i] - \tilde{\sigma}_{ik}^2 = (b_i - c_{ik} + \lambda_2)/b_i < 0$. Thus $f(x) = 0$ has a unique root $\hat{\sigma}_{ik}^2 = \tilde{\sigma}_{ik}^2/(1 + \lambda_2/b_i) \in (1, \tilde{\sigma}_{ik}^2)$.

$ii)$ When $\tilde{\sigma}_{ik}^2 < 1$, we have $b_i > c_{ik}$, and further $b_i - c_{ik} > \lambda_2$. Notice $f(x) = x[1 + \lambda_2/b_i] - \tilde{\sigma}_{ik}^2 = (x - \tilde{\sigma}_{ik}^2) + x\lambda_2/b_i > 0$ if $x > 1$. Thus the possible root of $f(x) = 0$ should be smaller than 1. For $x < 1$, $f(x) = x[1 - \lambda_2/b_i] - \tilde{\sigma}_{ik}^2$ is a linear function of $x$. $f(\tilde{\sigma}_{ik}^2) = -\lambda_2 \tilde{\sigma}_{ik}^2/b_i < 0$, and $\lim_{x \to 1^-} f(x) = [1 - \lambda_2/b_i] - \tilde{\sigma}_{ik}^2 = (b_i - c_{ik} - \lambda_2)/b_i > 0$. Thus $f(x) = 0$ has a unique root $\hat{\sigma}_{ik}^2 = \tilde{\sigma}_{ik}^2/(1 - \lambda_2/b_i) \in (\tilde{\sigma}_{ik}^2, 1)$.

Based on the signs of $f(x)$ around $x = \hat{\sigma}_{ik}^2$, it is easy to see that $\hat{\sigma}_{ik}^2$ is indeed a local maximizer. And $\hat{\sigma}_{ik}^2 = \tilde{\sigma}_{ik}^2/(1 + \text{sign}(c_{ik} - b_i)\lambda_2/b_i)$

### B.6. Derivation of Theorem 4

Consider two cases:

$i)$ $\mu_i^m \ne \mathbf{0}$. First, by definition and using the Cauchy-Schwarz inequality, we can prove that the $L_2$ norm is convex, thus the penalty function for grouped means is convex in $\mu_i^m$. Second, treating $Q_P$ as the Lagrange multiplier for a constrained optimization problem with the penalty as the inequality constraint, and considering that both minus the objective function and the penalty function are convex, by the Karush-Kuhn-Tucker (KKT) condition, we have the following sufficient and necessary condition

$$\partial Q_P/\partial \mu_i^m = \mathbf{0} \iff \sum_j \tau_{ij} V_{im}^{-1}(x_j^m - \mu_i^m) - \lambda_1 \sqrt{k_m} \mu_i^m/||\mu_i^m|| = \mathbf{0},$$

from which we can easily get (21).

ii) $\mu_i^m = \mathbf{0}$. By definition, we have

$$Q_P(\mathbf{0},.) \geq Q_P(\Delta\mu_i^m,.) \text{ for any } \Delta\mu_i^m \text{ close to } \mathbf{0}$$

$$\iff -\sum_j \tau_{ij}\frac{1}{2}(x_j^m)'V_{im}^{-1}x_j^m + C_1 \geq$$

$$-\sum_j \tau_{ij}\frac{1}{2}(x_j^m - \Delta\mu_i^m)'V_{im}^{-1}(x_j^m - \Delta\mu_i^m) - \lambda_1\sqrt{k_m}||\Delta\mu_i^m|| + C_1$$

$$\iff \sum_j \tau_{ij}x_j^{m\prime}V_{im}^{-1}\Delta\mu_i^m/||\Delta\mu_i^m|| - \sum_j \tau_{ij}(\Delta\mu_i^m)'V_{im}^{-1}\Delta\mu_i^m/(2||\Delta\mu_i^m||) \leq$$

$$\lambda_1\sqrt{k_m}.$$

Plugging-in $\Delta\mu_i^m = \alpha\sum_j \tau_{ij}V_{im}^{-1}x_j^m$ and letting $\alpha \to 0$, we obtain (22) from the above inequality. On the other hand, by the Cauchy-Schwarz inequality, we have $\left|\sum_j \tau_{ij}x_j^{m\prime}V_{im}^{-1}\Delta\mu_i^m/||\Delta\mu_i^m||\right| \leq ||\sum_j \tau_{ij}x_j^{m\prime}V_{im}^{-1}||$, and because $V_{im}^{-1}$ is positive definite, we obtain the above inequality from (22).

### B.7. Derivation of Theorem 5

If $\sigma_{i,m}^2 = \mathbf{1}$ is a local maximum, by definition, we have the following sufficient and necessary condition

$$Q_P(\mathbf{1},.) \geq Q_P(\mathbf{1} + \Delta\sigma_{i,m}^2,.) \text{ for any } \Delta\sigma_{i,m}^2 \text{ near } \mathbf{0}$$

$$\iff \sum_j \tau_{ij}\left[-\frac{1}{2}(x_j^m - \mu_i^m)'(x_j^m - \mu_i^m)\right] + C_1 \geq$$

$$\sum_j \tau_{ij}\left[-\frac{1}{2}\log|\text{diag}(\mathbf{1} + \Delta\sigma_{i,m}^2)|\right.$$

$$\left.-\frac{1}{2}(x_j^m - \mu_i^m)'\text{diag}(\mathbf{1} + \Delta\sigma_{i,m}^2)^{-1}(x_j^m - \mu_i^m)\right]$$

$$- \lambda_2\sqrt{k_m}||\Delta\sigma_{i,m}^2|| + C_1.$$

Thus,

$$\sum_j \tau_{ij}\left[-\frac{1}{2}\log|\text{diag}(\mathbf{1} + \Delta\sigma_{i,m}^2)|\right.$$

$$\left.+\frac{1}{2}(x_j^m - \mu_i^m)'\text{diag}(\Delta\sigma_{i,m}^2/(\mathbf{1} + \Delta\sigma_{i,m}^2))(x_j^m - \mu_i^m)\right]$$

$$\leq \lambda_2\sqrt{k_m}||\Delta\sigma_{i,m}^2||.$$

Using Taylor's expansion, we have

$$\sum_{j=1}^n \tau_{ij}\left(-\frac{1}{2}\mathbf{1} + \frac{(x_j^m - \mu_i^m)^2}{2}\right)'\Delta\sigma_{i,m}^2 + \frac{1}{2}\sum_{j=1}^n \tau_{ij}\left(\frac{1}{2}\mathbf{1} - (x_j^m - \mu_i^m)^2\right)'(\Delta\sigma_{i,m}^2)^2$$

$$+ O\left(\mathbf{c}'(\Delta\sigma_{i,m}^2)^3\right) \leq \lambda_2\sqrt{k_m}||\Delta\sigma_{i,m}^2||$$

for some constant vector **c**. After dividing both sides by $\|\Delta\sigma_{i,m}^2\|$ and using the same argument as before, we obtain (24) as the sufficient and necessary condition for $\sigma_{i,m}^2 = \mathbf{1}$ to be a local maximizer of $Q_P$.

### *B.8. Characterization of solutions to (25)*

Consider any component $k'$, $\sigma_{ik'}^2$, of $\sigma_{i,m}^2$. Equation (25) corresponds to

$$-\frac{b_{ik'}}{\sigma_{ik'}^2} + \frac{c_{imk'}}{(\sigma_{ik'}^2)^2} = a_{im}(\sigma_{ik'}^2 - 1),$$

where $a_{im} = \lambda_2\sqrt{k_m}/\|\sigma_{i,m}^2 - \mathbf{1}\|$, and $b_{ik'}$ and $c_{imk'}$ are the $k'$th components of $\mathbf{b_i}$ and $\mathbf{c_{im}}$ respectively. If $\lambda_2 = 0$, then $\sigma_{ik'}^2 = c_{imk'}/b_{ik'} = \tilde{\sigma}_{ik'}^2$, the usual MLE without penalization; if $\lambda_2 \neq 0$ and we treat $a_{im}$ as a constant (i.e. by plugging-in a current estimate of $\sigma_{ik'}^2$), the above equation becomes a cubic equation of $\sigma_{ik'}^2$, $f(\sigma_{ik'}^2)$,

$$f(x) = x^3 + ax^2 + bx + c$$

where $a = -1, b = b_{ik'}/a_{im}, c = -c_{imk'}/a_{im}$.

Now we consider the following two cases:

i) When $\tilde{\sigma}_{ik'}^2 = c_{imk'}/b_{ik'} > 1$, we have $f(\tilde{\sigma}_{ik'}^2) = (\tilde{\sigma}_{ik'}^2)^2(\tilde{\sigma}_{ik'}^2 - 1) > 0$, $f(1) = b(1 - \tilde{\sigma}_{ik'}^2) < 0$, $f(x) < 0$ for $\forall x < 1$, and $f(x) > 0$ for $\forall x > \tilde{\sigma}_{ik'}^2$. Therefore, the real roots of this equation must be between 1 and $\tilde{\sigma}_{ik'}^2$. Recall the fact that an odd-order equation has at least one real root, and the sum of all three roots of this equation equals $-a = 1$, the equation must have only one real root $\hat{\sigma}_{ik'}^2 \in (1, \tilde{\sigma}_{ik'}^2)$. Because $f(\sigma_{ik'}^2) = -\partial Q_P/\partial\sigma_{ik'}^2$, based on the signs of $f(x)$ near $x = \hat{\sigma}_{ik'}^2$, we know that $\hat{\sigma}_{ik'}^2$ is a local maximizer.

ii) When $\tilde{\sigma}_{ik'}^2 = c_{imk'}/b_{ik'} < 1$, we have $f(\tilde{\sigma}_{ik'}^2) = (\tilde{\sigma}_{ik'}^2)^2(\tilde{\sigma}_{ik'}^2 - 1) < 0$, $f(1) = b(1 - \tilde{\sigma}_{ik'}^2) > 0$, $f(x) > 0$ for $\forall x > 1$, and $f(x) < 0$ for $\forall x < \tilde{\sigma}_{ik'}^2$. Therefore, the real roots of this equation must be between $\tilde{\sigma}_{ik'}^2$ and 1. By factorization, we have

$$f(x) = x^3 + ax^2 + bx + c = (x - x_1)(x^2 + (a + x_1)x + b + ax_1 + x_1^2)$$

where $x_1$ is a root of $f(x) = 0$. Thus, if we use a bisection search to find the first root $x_1$, the other two (real or complex) roots of $f(x) = 0$ are

$$x_{2,3} = \left(-(a + x_1) \pm \sqrt{-3x_1^2 - 2ax_1 + a^2 - 4b}\right)/2.$$

If there is more than one real root, we choose the one maximizing $Q_P$ as the new estimate $\hat{\sigma}_{ik'}^2$.

### **Appendix C: Simulation**

### *C.1. Comparison of the two regularization schemes*

We investigated the performance of the two regularization schemes for the variance parameters for set-up 3 in simulation case I. There were 36 (or 5) out of

TABLE 8
*The mean numbers of the genes (or variables) whose penalized variance parameters were exactly one by the two regularization schemes, averaged over the datasets with $\hat{g} = 2$ and $\hat{g} = 3$ respectively. "Overlap" gives the common genes between the two regularization schemes or between/among the two/three clusters*

|   |             | $\hat{g}=2$ | | | $\hat{g}=3$ | | | |
|---|-------------|-----------|-----------|---------|-----------|-----------|-----------|---------|
|   |             | Cluster 1 | Cluster 2 | Overlap | Cluster 1 | Cluster 2 | Cluster 3 | Overlap |
|   | $L_1$(var-1) | 281.0 | 283.2 | 278.0 | 288.0 | 290.0 | 286.0 | 286.0 |
| 3 | $L_1$(logvar) | 277.3 | 279.2 | 272.5 | 288.0 | 290.0 | 286.0 | 286.0 |
|   | Overlap     | 276.8 | 278.9 | 271.9 | 288.0 | 290.0 | 286.0 | 286.0 |

100 datsets which were identified with 2 (or 3) clusters by both (var-1) and log(var) methods. Table 8 summarizes the numbers of the genes with their penalized variance estimates as exactly one by either regularization scheme. For $\hat{g} = 3$, the two schemes gave exactly the same number of the genes for each cluster and discovered the same genes with their variances estimated as one across all 3 clusters. For $\hat{g} = 2$, the results of the two schemes were also similar, though scheme one (i.e. penalizing var-1) identified slightly more genes with their variances estimated to be one for each cluster and more genes across both the clusters than did scheme two.

Figure 5 compares the variance parameter estimates by the two regularization schemes and the sample variance estimates based on the estimated sample assignments for the estimated $\hat{g} = 2$ clusters for one simulated dataset in set-up 3. Due to the construction of the simulation data and standardization, the true cluster with 80 samples always had sample variances smaller than 1 for informative variables, while the other cluster with 20 samples always had sample variances larger than 1 for those informative variables. Compared to the sample variance estimates, the penalized estimates from both schemes were clearly shrunken towards 1, and could be exactly 1. Between the two schemes, they gave similar estimates for cluster 2, but scheme 1 in general shrank many variance parameters more than scheme 2, which was in agreement with and explained the results in Table 8.

## Appendix D: Golub's data

### D.1. Comparison of the two regularization schemes

Figure 6 compares the MPLEs of the variance parameters given by the two regularization schemes for Golub's data with the top 2000 genes. Although the two schemes in general gave similar MPLEs, scheme 1 seemed to shrink more than did scheme 2, especially if $\hat{\sigma}^2 > 1$.

Figures 7–8 compare the MPLEs from the two schemes with the sample variances based on the final clusters. The effects of shrinkage to and thresholding at 1 by the two regularization schemes were striking. In particular, there was a clear thresholding in MPLE when the sample variances were less than and close to 1 for scheme 1 (Figure 7). To provide an explanation, we examined expression

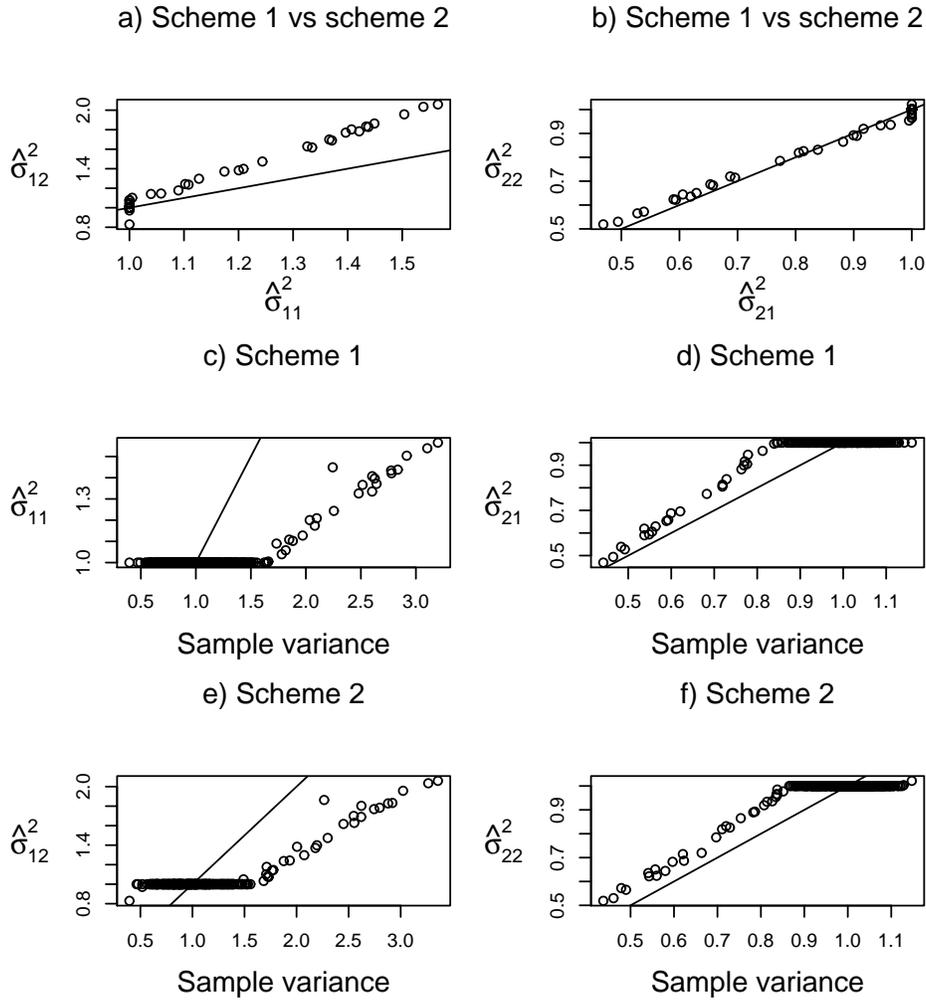

FIG 5. *Comparison of the two regularization schemes on the variance parameters for one dataset of set-up 3. $\hat{\sigma}_{is}$ is MPLE for cluster $i$ by scheme $s$.*

(15) given in the paper. We notice that if $\tilde{\sigma}_{ik}^2$ (in the form of the usual MLE) is less than 1, and $\lambda_2$ is large enough, then the MPLE $\hat{\sigma}_{ik}^2 < \tilde{\sigma}_{ik}^2/(1/2 + 0) = 2\tilde{\sigma}_{ik}^2 < 2(1 - \lambda_2/b_i) < 1$. Therefore, $\hat{\sigma}_{ik}^2$ did have a ceiling at $2(1 - \lambda_2/b_i)$.

### *D.2. Comparison with Kim et al. (2006)'s method*

We applied our penalized clustering methods to the Golub's data that were pre-processed as in Kim et al. (2006) [25], resulting in 3571 genes; see Table 9.

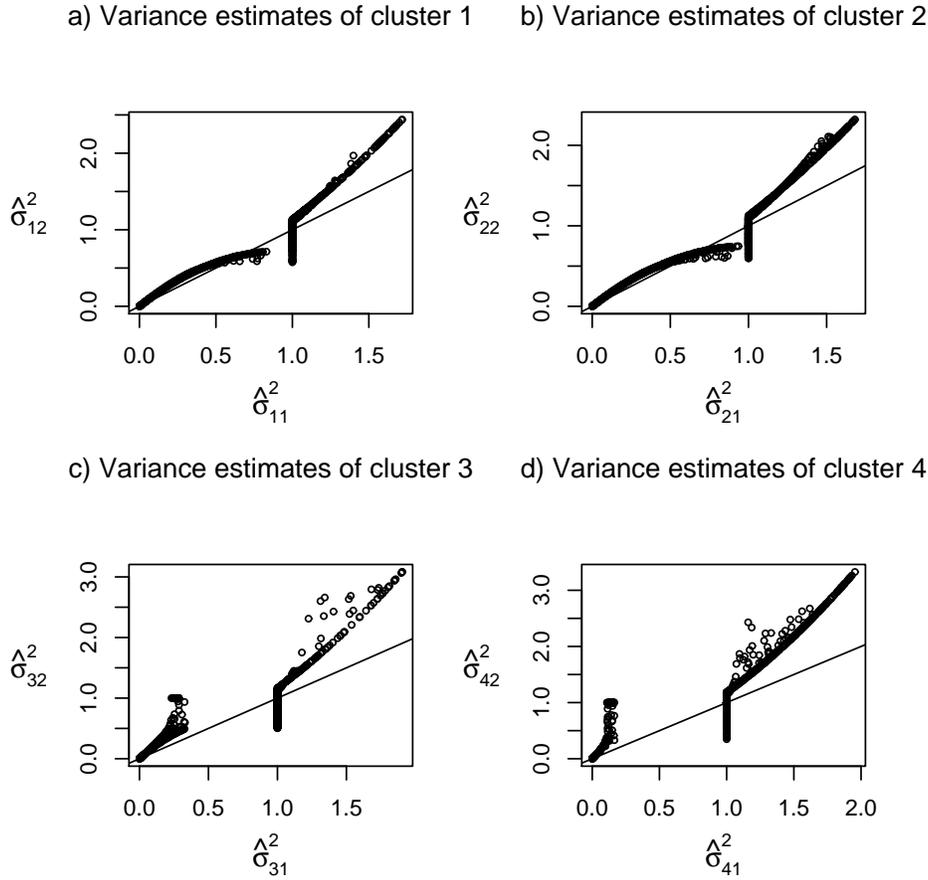

FIG 6. *Comparison of the two regularization schemes on the variance parameters for Golub's data with the top 2000 genes. X-axis and y-axis give the MPLEs by scheme 1 and scheme 2 respectively.*

The standard methods without variable selection under-selected the number of clusters at 2, failing to distinguish between ALL-T and ALL-B, even between ALL and AML (for the equal covariance model), in agreement with our simulation results. Our proposed new penalized method could largely separate the AML samples and the two ALL subtypes; only two samples were mis-assigned. In contrast, Kim *et al*'s method could not separate the two subtypes of ALL samples.

## Acknowledgements

We thank the editor for an extremely timely and helpful review. This research was partially supported by NIH grant GM081535; in addition, BX and WP by

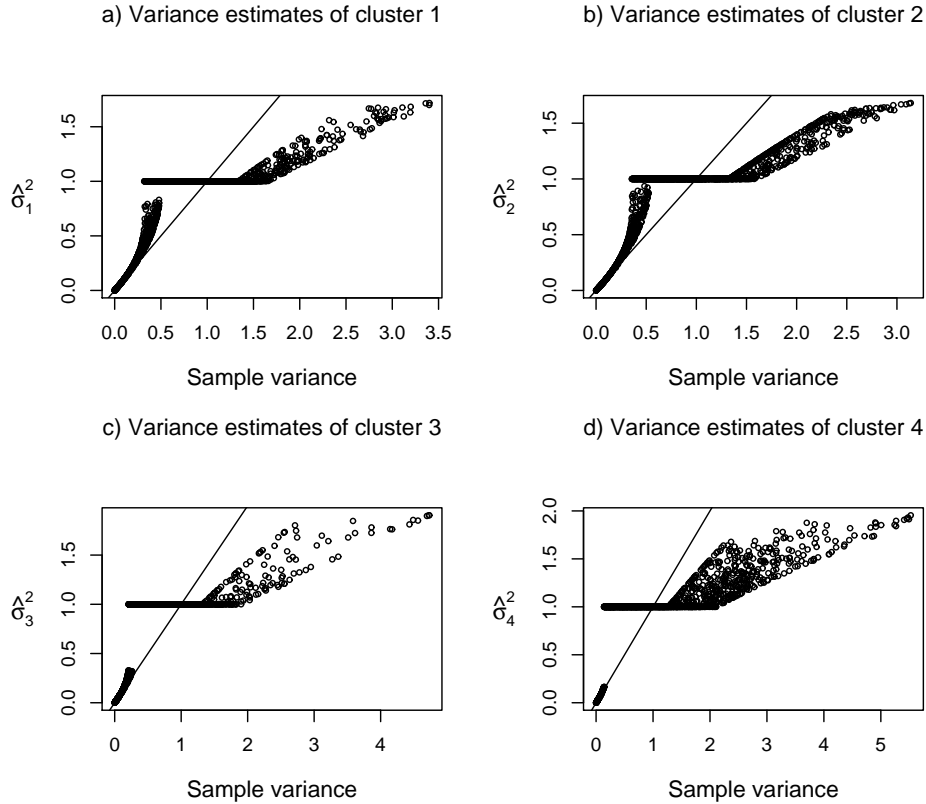

Fig 7. *Comparison of the penalized variance estimates by regularization scheme 1 and the sample variances for Golub's data with the top 2000 genes.*

Table 9
*Clustering results for Golub's data with 3571 genes. The number of components (g) was selected by BIC*

| Methods | UnequalCov($\lambda_1, \lambda_2$) | | | | | | | EqualCov($\lambda_1$) | | | | | |
|---|---|---|---|---|---|---|---|---|---|---|---|---|---|
| | (0, 0) | | ($\hat{\lambda}_1, \hat{\lambda}_2$) | | | | | (0) | | ($\hat{\lambda}_1$) | | | |
| BIC | 148898 | | 116040 | | | | | 134660 | | 124766 | | | |
| Clusters | 1 | 2 | 1 | 2 | 3 | 4 | 5 | 1 | 2 | 1 | 2 | 3 | 4 |
| Samples(#) | | | | | | | | | | | | | |
| ALL-T(8) | 0 | 8 | 0 | 0 | 8 | 0 | 0 | 5 | 3 | 0 | 0 | 8 | 0 |
| ALL-B(19) | 5 | 14 | 10 | 1 | 0 | 7 | 1 | 11 | 8 | 11 | 1 | 0 | 7 |
| AML(11) | 11 | 0 | 0 | 10 | 0 | 0 | 1 | 1 | 10 | 0 | 11 | 0 | 0 |

NIH grant HL65462 and a UM AHC Faculty Research Development grant, and XS by NSF grants IIS-0328802 and DMS-0604394.

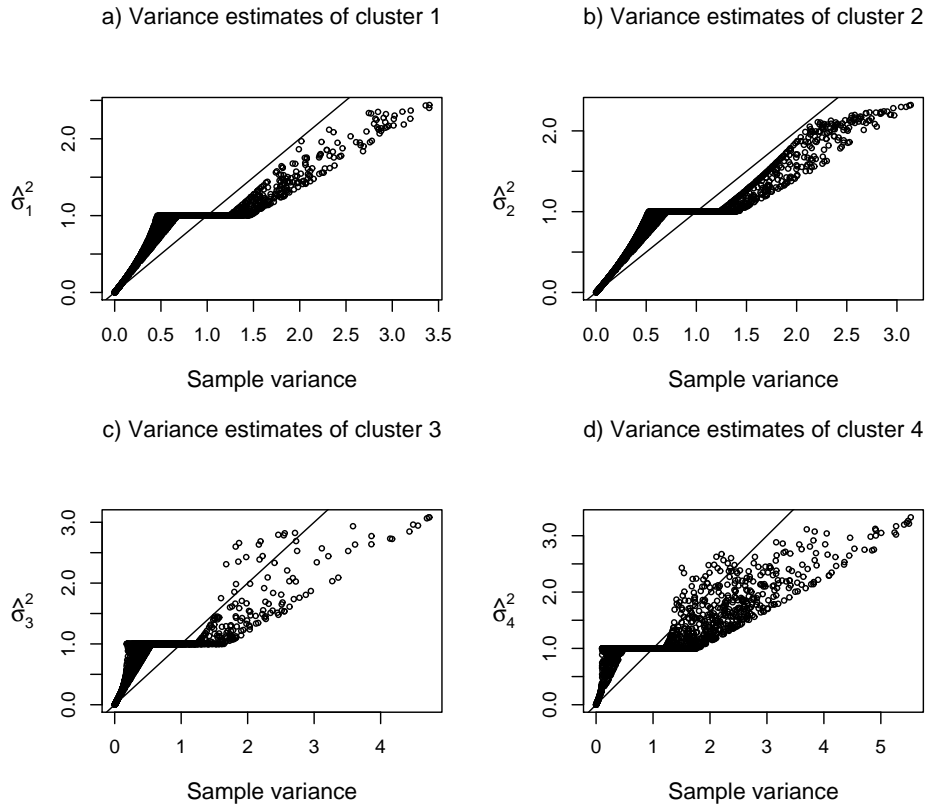

Fig 8. *Comparison of the penalized variance estimates by regularization scheme 2 and the sample variances for Golub's data with the top 2000 genes.*